# Estimation of minimum miscibility pressure (MMP) in impure/pure N$_2$ based enhanced oil recovery process: A comparative study of statistical and machine learning algorithms

Xiuli Zhu[1], Seshu Kumar Damarla[2], Biao Huang[2][*]

[1] University of Shanghai for Science and Technology, China

[2] Department of Chemical and Materials Engineering, University of Alberta, Edmonton, Alberta, T6G2V4, Canada

**Abstract:** MMP prediction plays an important role in design and operation of nitrogen based enhanced oil recovery processes. In this work, a comparative study of statistical and machine learning methods used for MMP estimation is carried out. Most of the predictive models developed in this study exhibited superior performance over correlation and predictive models reported in literature.

## 1. Introduction

Crude oil from underground oil reservoirs is extracted using conventional primary and secondary recovery methods. In the course of primary recovery, oil is pushed towards a production well by the natural pressure of reservoir, and then it is brought to surface of the earth by artificial lift devices such as pump jacks. Only 10 percent of reservoir's original oil in place (OOIP) is recovered in this phase. As the reservoir pressure uses up, oil production begins to decline that marks the end of the primary recovery phase. At this point, the secondary recovery phase commences with injection of water into the reservoir. The injected water maintains reservoir pressure required to continue to displace and drive the oil to the production well, resulting in additional recovery of 20-40 percent of OOIP. For the entire duration of the primary and secondary recovery phases, properties of the reservoir oil do not change but the oil is merely driven, in the direction of the production well, by the reservoir pressure. The primary and secondary recovery methods can together recover utmost 40 percent of OOIP, deserting the remaining (i.e. residual oil) [1]. Prior to the development of tertiary recovery method (popularly known as enhanced oil recovery (EOR)), reservoirs with considerable amount of residual oil used to be abandoned after the primary and secondary recovery methods had been applied.

[*] Corresponding author. E-mail address: bhuang@ualberta.ca

Ever since EOR methods were first successfully tested and commercially applied to Kelly-Snyder oil field in the US, they have been widely employed to recover most of the residual oil in depleted (mature) reservoirs all over the world [2-5].

Among the existing EOR methods, thermal recovery, microbial injection, chemical injection and gas injection (gas flooding) were found to be commercially successful. Nevertheless, the gas injection process has received the greatest attention owing to the facts that it can upsurge oil recovery by mobilizing oil trapped in porous rocks, and the associated operational cost is noticeably less compared to the other methods. Gases often used in the gas injection process include carbon dioxide ($CO_2$), nitrogen ($N_2$), flue gas, natural gas and methane. Among the gas injection processes, $CO_2$ has been acknowledged as the widely preferred gas in EOR operations to recover residual oil from mature reservoirs containing light and medium oils. However, the use of $CO_2$ causes disadvantages like asphaltene precipitation, corrosion in production well and surface facilities, etc. The pros of the $CO_2$ based EOR process prompted oil operators to use $N_2$ based EOR process. Cantarell-Akal filed in Mexico's Bay of Campeche has currently the world's largest $N_2$ injection project in operation.

Once a reservoir is flooded with $N_2$, multiple-contact miscibility between the light components of oil and gas transpires. As miscibility improves, interfacial tension (IFT) keeps on plummeting and in due course reaches zero. At this stage, $N_2$ entirely dissolves in the oil and establishes a single phase. As $N_2$ dissolves in the oil, it swells oil and decreases oil viscosity, which improves flow capacity of the oil. At the same time, oil stuck in porous rocks as well as in between fine rocks emerges and moves toward the production well. Recovered residual oil contains some of the injected $N_2$, which is separated and reused. An additional 5-15 percent of OOIP can be recovered with a successful $N_2$ based EOR project. $N_2$ based EOR can be implemented using schemes like continuous gas injection, simultaneous gas injection, water alternate gas injection and hybrid water alternate gas injection. $N_2$ cannot mix with oil below a minimum pressure, which is called minimum miscibility pressure (MMP). When reservoir pressure is above MMP, both N2 and oil exhibit similar properties and dissolve into each other. So IFT between N2 and oil no longer exists, developing multiple-contact miscibility. If MMP is not reachable in any mature oil reservoir, N2 cannot blend with oil so that miscibility never happens, which leads to a great reduction in oil production.

Accurate determination of MMP plays a crucial role in designing $N_2$ based EOR process before conducting field EOR operation. Methods used for determining MMP in the $N_2$-crude oil system can be categorized into three groups: experimental methods, empirical correlations and predictive modelling techniques. The experimental methods mainly consist of slim tube test, vanishing interfacial tension method and rising bubble apparatus method. A detailed review of various experimental methods utilized for measuring MMP is reported elsewhere [6]. Despite being able to provide accurate measurements, these methods are not cost-effective and require more than a week to yield a single measurement. To obtain MMP more quickly, many empirical correlations were developed based on the data produced by the experimental methods [7-12]. Albeit the empirical correlations are inexpensive, they lack generality i.e. a correlation developed for a specific oil reservoir may not be applicable to another oil reservoir. This is because oil composition, temperature and pressure vary among different reservoirs. In addition, MMP estimations produced by the correlations are not highly accurate. The limitations of the experimental methods and the empirical correlations prompted researchers to resort to the predictive modelling techniques.

Machine learning (ML) and artificial intelligence benefits oil and gas industries by providing systems the ability to automatically learn and improve from experience without being explicitly programmed. These ubiquitous techniques are being utilized in many energy industries to improve process operations in varied fields such as drilling operations, subsurface characterization, process monitoring, inferential measurement, reservoir modelling, predictive maintenance, etc. [13-15]. The first application of ML algorithms in the design of $N_2$ based EOR project was reported in Hemmati-Sarapardeh et al. (2016) [16]. In [16], the authors developed least squares support vector machine (LSSVM) based prediction model, whose parameters were optimized by coupled simulated annealing, using database spanning an extensive range of operational parameters collected from reliable literature. The LSSVM model was used to estimate MMP in pure and impure $N_2$-crude oil system. In [17], the authors carried out a comparative study of three ML algorithms (multilayer perceptron (MLP), radial basis neural network (RBFNN) and adaptive neuro-fuzzy inference system (ANFIS)) based on the same database that was used in [16]. The authors found that RBFNN model is capable of providing more accurate estimations for MMP than the remaining two. The prediction models reported in [16, 17] were found to be better than the empirical correlations derived in [7-12]. Quest for better predictive models based on ML or statistical learning (SL) algorithms continues until highest possible accuracy is achieved.

**1.2. Motivation and contributions**

Research gaps in the previous works are identified as below.

- It is noticed that training (or calibration) dataset was created by randomly picking samples from data historian. However, there is no guarantee that the training dataset generated, in this fashion, include samples falling near or on the boundaries of input space of given labelled data. Thus, the prediction models may suffer from overfitting.

- The prediction models are mainly based on artificial neural networks (ANNs) and LSSVM. Nevertheless, there exist much simpler supervised SL and ML algorithms, which can demonstration analogous and/or better prediction performance than ANNs and SVM. Some of these methods require no or less training.
- It is also noticed that some elegant supervised methods, which can estimate MMP with reasonable accuracy, are totally ignored by the research fraternity.

The main contributions of the present work are summarized below.
- Data partition algorithm: HSPXY is adopted in the present study to prepare training and testing datasets from the database of pure/impure $N_2$-crude oil MMP.
- Various classes of SL and ML methods are applied to the MMP estimation problem. These methods include linear regression models (multiple linear regression and polynomial regression), feed-forward ANNs (extreme learning machine, general regression neural network and wavelet neural network), nearest neighbors method (k-nearest neighbors for regression), graphical methods (regression trees and associated ensemble methods) and kernel methods (relevance vector machine and Gaussian process regression).
- Sensitivity of the best prediction model obtained from the above step is conducted to examine the influence of each of the input variables on MMP.

The rest of the manuscript is organized as follows. The theory of the chosen SL and ML methods is given in Section 2. In Section3, statistical analysis of the practical data used for the building predictive model is carried out, and a data partition algorithm (HSPXY) is introduced to split the database into training and testing datasets. In Section 5, the construction of each prediction model is explained in detail, and comparative and sensitivity analysis are also presented. Finally, conclusions drawn from the present comparative study are provided.

## 2. Preliminaries

In this section, a brief overview and tutorial of SL and ML methods used in this paper is provided.

### 2.1. Linear regression methods

#### 2.1.1 Multiple linear regression

Multiple linear regression (MLR) is the simplest statistical method that relates MMP to the input variables via the linear model given below [18].

$$y_i = \beta_0 + \beta_1 x_{1i} + \beta_2 x_{2i} + \cdots + \beta_P x_{Pi}, \quad i = 1, 2, \ldots, N \tag{1}$$

where $\beta_j$'s are unknow model parameters, $x_{ji}$ is the $i^{th}$ training observation of the $j^{th}$ input variable, $N$ is the number of training samples and $P$ is the number of input variables. Eq. (1) can be rewritten in the matrix notation shown below.

$$\begin{bmatrix} y_1 \\ y_2 \\ y_3 \\ \vdots \\ y_N \end{bmatrix} = \begin{bmatrix} 1 & x_{11} & x_{21} & \cdots & x_{P1} \\ 1 & x_{12} & x_{22} & \cdots & x_{P2} \\ 1 & x_{12} & x_{22} & \ddots & x_{P2} \\ \vdots & \vdots & \vdots & \ddots & \vdots \\ 1 & x_{1N} & x_{2N} & \cdots & x_{PN} \end{bmatrix} \begin{bmatrix} \beta_1 \\ \beta_2 \\ \beta_3 \\ \vdots \\ \beta_N \end{bmatrix} \quad (2)$$

$$\underbrace{\phantom{\begin{bmatrix} y_1 \end{bmatrix}}}_{Y} \quad \underbrace{\phantom{\begin{bmatrix} 1 \end{bmatrix}}}_{X} \quad \underbrace{\phantom{\begin{bmatrix} \beta \end{bmatrix}}}_{\beta}$$

The determination of the unknown parameter vector $\beta$ can be expressed as an optimization problem:

$$J = \underset{\beta}{\arg\min} \sum_{i=1}^{N} (y_i - \tilde{y}_i)^2, \quad (3)$$

where $\tilde{y}_i$ is the approximation for $y_i$.

The solution of the above optimization problem can be obtained as shown in Eq. (4) via the method of least squares.

$$\beta = \left(X^T X\right)^{-1} X^T Y. \quad (4)$$

### 2.1.2. Polynomial regression (PR)

Polynomial regression (PR) is an extension of MLR and includes linear and quadratic term for each input variable, and all products of pairs of distinct input variables [18].

$$y_i = \beta_0 + \sum_{j=1}^{P} \beta_j x_{ji} + \sum_{k=1}^{P} \beta_{P+k} x_{ki}^2 + \sum_{m=2}^{P} \beta_{2P+m-1} x_{1i} x_{mi} + \sum_{q=3}^{P} \beta_{(3P-1)+q-2} x_{2i} x_{qi} + A, \quad i=1,2,\ldots,N \quad (5)$$

where $A = \beta_{(4P-2)} x_{3i} x_{4i} + \beta_{(4P-1)} x_{3i} x_{5i} + \beta_{4P} x_{4i} x_{5i}$.

The next equation provides a concise representation of Eq. (5).

$$\begin{bmatrix} y_1 \\ \vdots \\ \vdots \\ y_q \\ \vdots \\ \vdots \\ y_N \end{bmatrix} = \begin{bmatrix} 1 & x_{11} & \cdots & x_{11}^2 & \cdots & x_{11} x_{21} & \cdots & x_{(P-1)1} x_{p1} \\ 1 & x_{12} & \cdots & x_{12}^2 & \cdots & x_{12} x_{22} & \cdots & x_{(P-1)2} x_{p2} \\ 1 & x_{13} & \cdots & x_{13}^2 & \cdots & x_{13} x_{23} & \cdots & x_{(P-1)3} x_{p3} \\ \vdots & \vdots & \vdots & \vdots & \vdots & \vdots & \vdots & \vdots \\ 1 & x_{1q} & \cdots & x_{1q}^2 & \cdots & x_{1q} x_{2q} & \cdots & x_{(P-1)q} x_{pq} \\ \vdots & \vdots & \vdots & \vdots & \vdots & \ddots & \vdots & \vdots \\ \vdots & \cdots & \cdots & \cdots & \cdots & \cdots & \ddots & \vdots \\ 1 & x_{1N} & \cdots & x_{1N}^2 & \cdots & x_{1N} x_{2N} & \cdots & x_{(P-1)2} x_{p2} \end{bmatrix} \begin{bmatrix} \beta_0 \\ \vdots \\ \beta_P \\ \vdots \\ \beta_{2P} \\ \vdots \\ \beta_{4P} \end{bmatrix} \quad (6)$$

Eq. (6) is an overdetermined linear system (more equations than unknowns), hence, the method of least squares can be employed to minimize the residual sum of squares to obtain optimal $\beta$ ($\beta = \beta_0, \ldots, \beta_{4P}$).

## 2.2. Weighted K-nearest neighbors for regression

K-nearest neighbors (KNN) method is one of the simplest non-parametric ML method used for regression, classification and anomaly detection [19]. KNN does not involve explicit training but storage of input observations and corresponding response values of the training samples. There are three main steps embroiled in estimating MMP using the KNN regression:

- Distance between a given test (or query) sample and each of the training samples is computed.
- Based on the computed distances, k training samples that are closest to the test sample are selected.
- MMP of the test sample is obtained by averaging the MMP values of the k nearest neighbors (training samples).

The weighted KNN (WKNN) provides improved predictions by taking into consideration distance between the test sample and each of the k-nearest neighbors. The response value of each of the k-nearest neighbors is multiplied by a weight that is equal to inverse of the squared distance from that neighbor to the test sample.

$$\tilde{y} = \frac{\sum_{j=1}^{k}\left(y_j / w_j\right)}{k}, \qquad (7)$$

where $w_j = \|x_j - \hat{x}\|_2^2$ and $\hat{x}$ is the given test sample.

Euclidean distance is the preferred distance metric for continuous input variables. The performance of WKNN depends on the selection of k. In the present work, 5-fold cross validation approach is adopted to determine optimal k. The procedure of determining the optimum number of nearest neighbors is explained in Fig. 1.

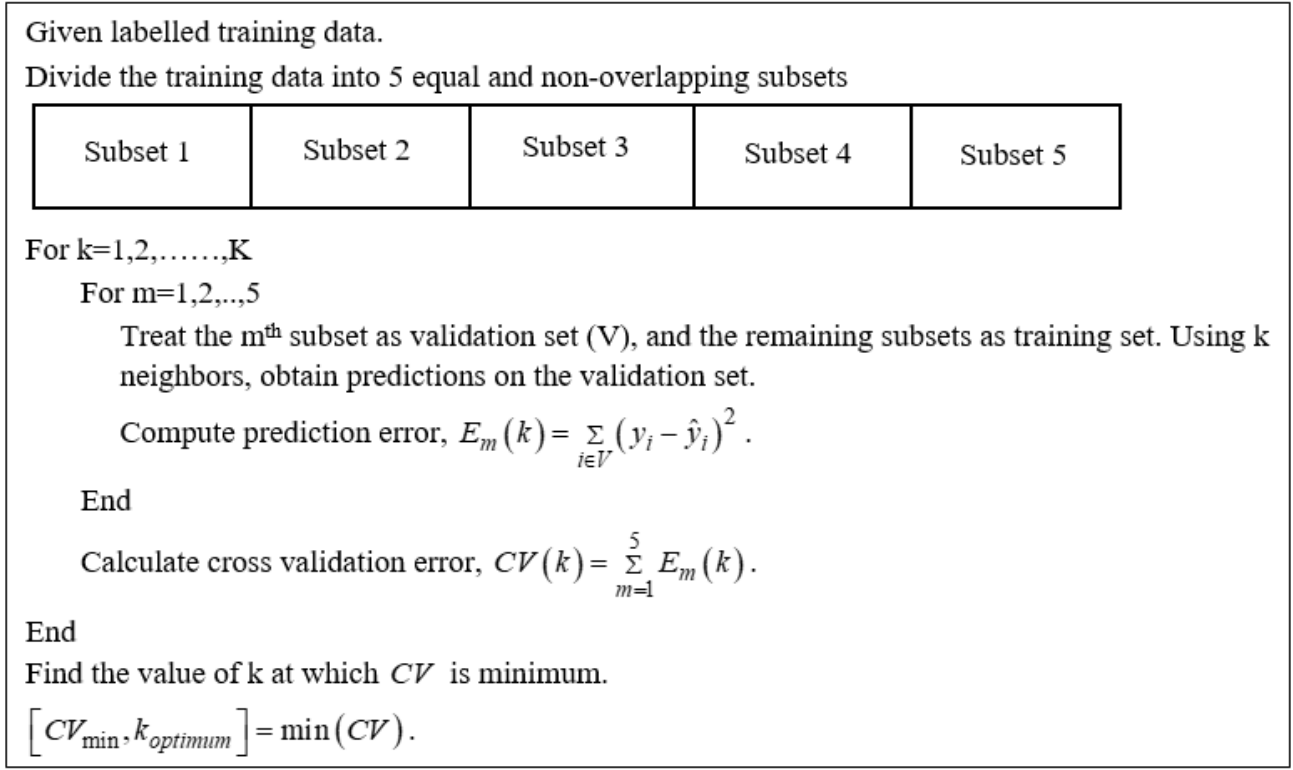

**Figure 1.** 5-fold cross validation for optimum k.

## 2.3. ANNs

### 2.3.1. Wavelet neural network (WNN)

Wavelet neural network (WNN) proposed by Zhang and Benveniste (1992) is a hybrid of wavelet transform (WT) and multilayer perceptron (MLP) [20]. WNN inherits the strengths of WT and MLP. The configuration of WNN is akin to that of MLP, and shown in Fig. 2. The input layer links the network to its environment i.e. estimating MMP in the $N_2$ based EOR process. The size of the input layer is determined by the number of input variables of the training dataset $\{x_i, y_i\}_{i=1}^{N}$ where $x_i = [x_{i1}, x_{i2}, ..., x_{iP}]$ and $y_i$ is a scalar response value for the corresponding $x_i$. The input layer is made up of source nodes (or sensory units) which receive and present the input vector of a training sample to the network. The input vector propagates through the hidden layer on a neuron-by-neuron basis and reaches the output neuron in the output layer. Each layer is connected to the next layer though synaptic weights. Input to each neuron in either the hidden layer or the output layer is a function of the outputs of the neurons in the previous layer and synaptic weights associated with that neuron.

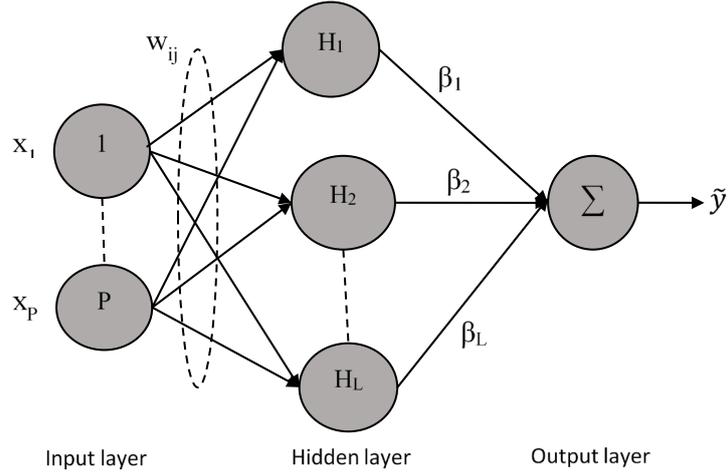

**Figure 2.** Structure of WNN.

Unlike in MLP, the hidden layer neurons use wavelets as activation function. The output of the $j^{th}$ hidden neuron is defined as

$$H(j) = H_j \left( \frac{\sum_{i=1}^{P} w_{ij} x_{ki} - b_j}{a_j} \right), \quad j = 1, 2, \ldots, L. \tag{8}$$

where $H_j$ is a mother wavelet function with the translation factor $b_j$ and the dilation factor $a_j$, $w_{ij}$'s are the weights connecting the input neurons with the $j^{th}$ hidden neuron, and $x_k$ is the $k^{th}$ input vector.

B-spline wavelet presented in Eq. (9) is considered as the mother wavelet function.

$$H_j = \sqrt{F_b} \operatorname{sinc}\left(\left(\frac{F_b x}{m}\right)^m\right) e^{2i\pi F_c x}, \text{ where } \operatorname{sinc}(b) = \begin{cases} 1 & \text{if } b = 0 \\ \frac{\sin(b)}{b} & \text{otherwise} \end{cases} \tag{9}$$

The output of WNN is computed as

$$\tilde{y} = \sum_{j=1}^{L} \beta_j H(j), \tag{10}$$

where $\beta_j$'s are weights of the output layer and $H(j)$ is the output of the $j^{th}$ hidden neuron.

Training algorithms used for MLP are equally applicable to WNN [21]. In this work, gradient descent algorithm is employed to adjust network parameters (weights, translation and dilation factors) in an online fashion.

The gradient descent algorithm iteratively chooses values for the unknown parameters in direction of the negative gradient of the loss function:

$$e_i = \frac{1}{2}(y_i - \tilde{y}_i)^2 \tag{11}$$

The rules for parameter updation are given below.

$$^{new}\beta_j = {}^{old}\beta_j - \eta_1\left(-(y_i - \tilde{y}_i)\right)H_j, \quad j = 1, 2, \ldots, L, \tag{12}$$

$$^{new}w^i_{jk} = {}^{old}w^i_{jk} - \eta_1\left[-(y_i - \tilde{y}_i)\frac{\partial H_j}{\partial input_j}\frac{x(k)}{a_j}\beta_j\right], \quad k = 1, 2, \ldots, P, \tag{13}$$

$$^{new}b_j = {}^{old}b_j - \eta_2\left[-(y_i - \tilde{y}_i)\beta_j\frac{\partial H_j}{\partial input_j}\left(\frac{-1}{a_j}\right)\right], \tag{14}$$

$$^{new}a_j = {}^{old}a_j - \eta_2\left[-(y_i - \tilde{y}_i)\beta_j\frac{\partial H_j}{\partial input_j}(net_j - b_j)\left(\frac{-1}{a_j^2}\right)\right], \tag{15}$$

where $input_j$ is the net input to the mother wavelet function $H_j$, $net_j = \sum_{k=1}^{P} wjk \times x_{ik}$ and $\eta_1$ and $\eta_2$ are the learning rates.

**Overview of WNN training**

**Normalization:** The input variables and the target variable are normalized to be in the range [-1, 1] using min-max normalization method.

**Initialization**: The network parameters are randomly initialized.

**Presentation of training observations**: In online mode of the gradient descent algorithm, in one epoch, all the training samples are presented to the neural network one at a time.

**Forward computation**: During this phase, the network parameters are not modified. The output of each hidden neuron is computed and then the output of the network is calculated. The error signal in Eq. (11) is calculated. The error signal propagates backward through the hidden layer to the input layer.

**Backward computation**: In this phase, in order to reduce deviation between the network output and the respective value of MMP, $\beta_j$'s are first adjusted, and then $w^i_{jk}$'s, $b_j$'s and $a_j$'s are changed using the rules given in Eqs. (12) through (15).

**Iteration**: Both the forward and backward computation phases are repeated for each training sample. Once an epoch is finished, the sum of square of the errors (SSE) is computed. The training is continued until maximum number of epochs (MaxE) is reached or SSE reaches a predefined threshold.

**Optimum number of hidden neurons**: The generalization performance of WNN also depends upon the size of the hidden layer. In this work, Widrow's rule of thumb given in Eq. (16) is used to determine the optimum size of the hidden layer [21]. For the given size ($N$) of the training set and effectiveness ($\varepsilon$) of the WNN model on test set, the number of free (or unknown) parameters in WNN

can be determined as

$$W = N \times \varepsilon.  \qquad (16)$$

**2.3.2. General regression neural network (GRNN)**

GRNN shown in Fig. 3 is a variant of radial basis function neural network (RBFNN) and a powerful tool for nonlinear function approximation. GRNN was developed by Specht (1991) based on idea that each training observation is assigned a neuron in the first hidden layer called pattern layer [22]. Compared to RBFNN, GRNN has one additional hidden layer called summation layer. There are no input weights connecting the input layer nodes to the pattern layer neurons. The traditional GRNN assumes that the pattern layer can have as many neurons as the given number of training examples. The pattern layer neurons use radial basis function (RBF) as activation function, thus these neurons are called radial basis neurons. The first training observation is selected as the center of the first radial basis neuron, the second training observation is the center of the second radial basis neuron and so on. All the radial basis neurons have the same width or spread ($\sigma_1 = \sigma_2 = \cdots \sigma_L = \sigma$).

The pattern layer is connected to the summation layer by the weights; $A^i$'s and $B^i$'s have the expression given in the equation below.

$$A^i = y_i, \quad B^i = 1, \quad i = 1, 2, \ldots, L, \qquad (17)$$

where $y_i$ is the actual output of the $i^{th}$ input vector $x_i = [x_1, x_2, \ldots, x_P]^T$.

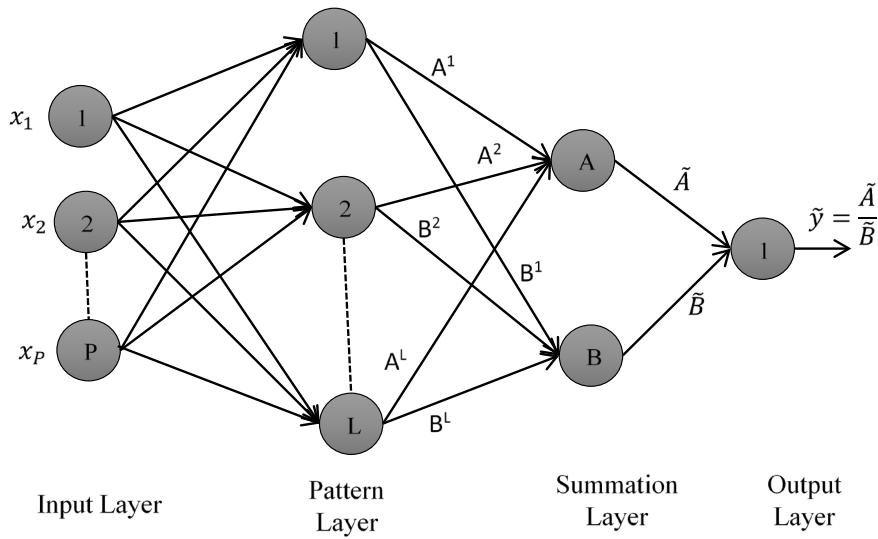

**Figure 3.** General regression neural network architecture

The summation layer has two neurons (A and B) that receive net input as the weighted sum of outputs of the pattern layer neurons:

$$U_A = \sum_{j=1}^{L} A^j \phi_j, \quad U_B = \sum_{j=1}^{L} B^j \phi_j. \qquad (18)$$

The linear activation function is employed in the neurons A and B. The outputs of A and B are denoted by $\tilde{A}$ and $\tilde{B}$, respectively.

$$\tilde{A} = U_A, \quad \tilde{B} = U_B. \tag{19}$$

The output layer has only one neuron that merely divides $\tilde{A}$ by $\tilde{B}$ to yield estimate for $y_i$. As GRNN has only one free parameter (spread) to be adjusted, its success highly depends on the value selected for $\sigma$. GRNN with very small $\sigma$ exactly learns from the training dataset with zero model errors but exhibits poor generalization performance. If a large value is selected for $\sigma$, model estimations assume a constant value which is simply the mean of the actual output values, therefore GRNN cannot learn the underlying relationship between the model inputs and the output so it cannot predict the new unseen samples. Therefore, the spread needs to be selected with great care. In the present work, a grid of candidate values for $\sigma$ is generated and the 5-fold cross validation approach is utilized to find optimum value for $\sigma$.

### 2.3.3. Extreme learning machine (ELM)

ELM is an extremely fast learning algorithm proposed by Huang et al. (2006) to train a single-hidden layer feedforward neural network (SLFN) shown in Fig. 4 [23].

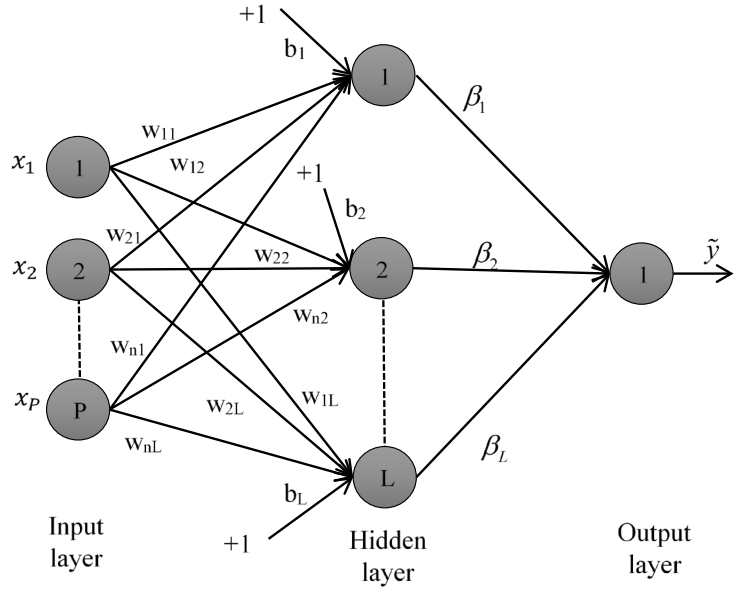

**Figure 4.** Architecture of SLFN.

In ELM, the input weights ($w_{11},\ldots,w_{1L}; w_{21},\ldots,w_{2L};\cdots;w_{k1},\ldots,w_{nL}$) and the biases ($b_1, b_2,\ldots,b_L$) are not learned iteratively; instead they are chosen randomly. The use of infinitely differentiable activation functions (in a given interval) such as sigmoid, RBF, sine, cosine, hyperbolic tangent function etc. in the hidden layer neurons actually allows ELM to assign random values to the input weights and biases. Once the input weights and biases are selected, SLFN becomes a linear system, hence, the output weights ($\beta = [\beta_1, \beta_1, \ldots, \beta_L]$) can be analytically determined. The mathematical theory of ELM is provided in the following.

The output of SLFN with $L$ hidden neurons and activation function $g(x)$ is obtained as

$$O_j = \sum_{i=1}^{L} \beta_i g_i (w_i \cdot x_j + b_i), \quad j = 1, 2, \ldots, N \tag{20}$$

where $w_i = [w_{i1}, w_{i2}, \ldots, w_{in}]^T$ is the weight vector connecting the $i^{th}$ hidden neuron to the input neurons, $\beta_i$ is a scalar weight connecting the output neuron to the $i^{th}$ hidden neuron, $x_j$ is the input vector of the $j^{th}$ training sample and $b_i$ is the bias of the $i^{th}$ hidden neuron.

In order to get zero model errors during learning, the output of SLFN is made equal to the actual output $y$ i.e.

$$\sum_{i=1}^{L} \beta_i g_i (w_i \cdot x_j + b_i) = y_j, \quad j = 1, 2, \ldots, N. \tag{21}$$

The above equations can be written in a matrix-vector form as shown below.

$$H\beta = T, \tag{22}$$

where $\beta = [\beta_1, \beta_2, \ldots, \beta_L]^T$, $T = [y_1, y_2, \ldots, y_N]^T$,

$$H = \begin{bmatrix} g(w_1 \cdot x_1 + b_1) & \cdots & g(w_L \cdot x_1 + b_L) \\ \vdots & \cdots & \vdots \\ g(w_1 \cdot x_N + b_1) & \cdots & g(w_L \cdot x_N + b_L) \end{bmatrix}.$$

The matrix $H$ is called the hidden layer output matrix of SLFN, and its $i^{th}$ column is the $i^{th}$ hidden neuron output with respect to the training samples. As the input layer weights and hidden layer biases are chosen randomly in the beginning of learning, the hidden layer output matrix is fixed. Therefore, the only unknown in Eq. (22) is the output weight vector $\beta$.

If the number of hidden layer neurons equals the number of given training samples ($L = N$), the matrix $H$ is invertible. Under this scenario, the output weights can be analytically determined from the following equation.

$$\beta = H^{-1} T. \tag{23}$$

The above solution is achievable when the number of training samples is considerably small. However, if the number of hidden layer neurons is significantly large, then the size of hidden layer output matrix $H$ is extremely big and requires huge storage space. In most applications of SLFN,

$L$ is usually selected to be much less than $N$. When $L \ll N$, the matrix $H$ becomes non-square and non-invertible. Therefore, Moore-Penrose generalized inverse is used to compute $\beta$:

$$\beta = H^{\dagger} T, \qquad (24)$$

where $H^{\dagger}$ is the Moore-Penrose generalized inverse of $H$.

## 2.4. Decision trees
### 2.4.1. Regression tree (RT)

A simple regression tree dividing a two-dimensional predictor space ($X = [X_1 \; X_2]$) into five separate regions ($R_1$, $R_2$ …, $R_5$) is displayed in Fig. 5 [18]. The regression tree is grown using the training data with the root node at its top and leaves at its bottom. In general, regression trees or classification trees are grown upside down to effortlessly understand the construction of regression or classification model. The regression tree is a binary tree since every time a split is made, the corresponding region of the predictor space gets divided into two smaller regions only. Splitting the predictor space begins at the top of the tree i.e. at the root node. The root node consists of all training observations, and has no parent. The root node is parent to child nodes 1 and 2. Leaf nodes 1 and 2 are the children of child node 1, and child node 2 is parent to leaf node 3 and child node 3 that is parent to leaf nodes 4 and 5. The child nodes are also called internal nodes while the leaf nodes are considered as terminal nodes. Each terminal node represents a distinct region in the predictor space. During the construction of a decision tree, only root node and internal nodes are split and terminal nodes do not undergo partition. An arrow connecting any two nodes is called a branch of the tree.

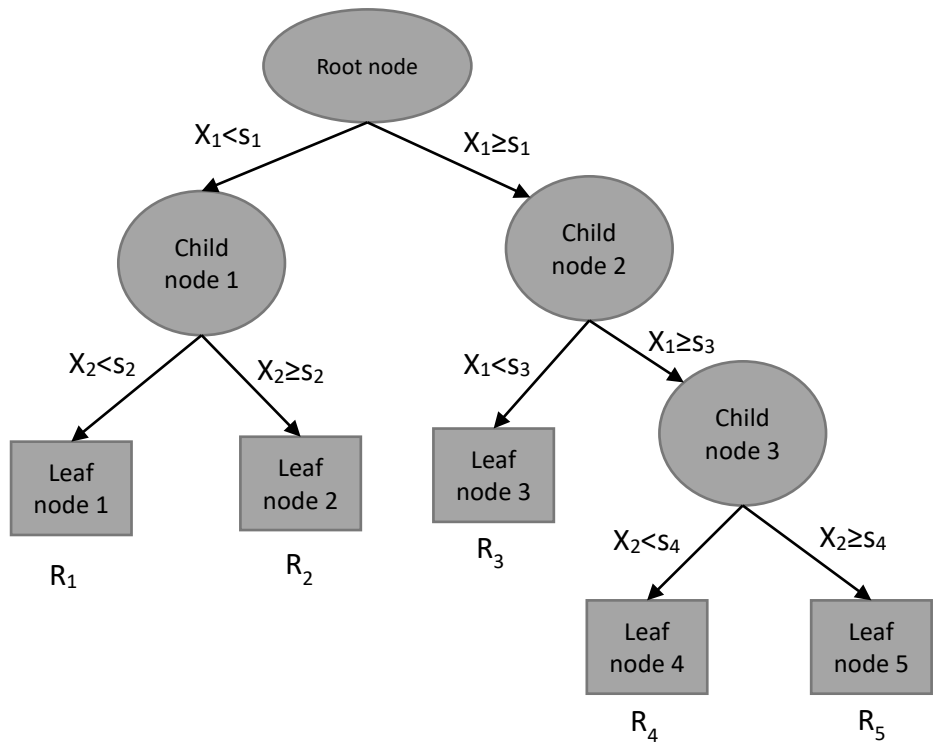

**Figure 5**. Regression tree. Here, $s_t$'s are cut-points dividing respective regions into two smaller regions.

The main purpose of the regression tree is to successively split the predictor space (for instance $X_1, X_2, ..., X_p$) into individual and non-overlapping regions $R_1, R_2, ..., R_J$, which minimize the residual sum of squares (RSS):

$$\text{RSS} = \sum_{j=1}^{J} \sum_{i \in R_j} \left( y_i - \hat{y}_{R_j} \right)^2, \qquad (25)$$

where $y_i$ is the actual response value for the $i^{th}$ training observation, $\hat{y}_{R_j}$ is the mean response for training observations that belong to $R_j$.

To perform partition, recursive binary splitting algorithm is adopted. The recursive binary splitting algorithm is a top-down approach as it carries out initial partition at the root node and then consecutively splits previously formed regions till no internal node is left to be divided into two parts. The recursive binary splitting algorithm first considers all predictors and all possible values of the cut-point $s$ for each of the predictors and then chooses the predictor and cut-point which give minimum RSS when the root node is divided into the left child node (internal node 1) and the right child node (internal node 2). The way of picking the best predictor and the best split is called greedy approach because only the predictor and its respective cut-point that give the lowest RSS when the current node is divided are selected, instead of choosing a predictor and its corresponding best cut-point that lead to a better tree when splitting a future node. The initial split is described mathematically in the following equation.

$$(j, s) = \underset{j,s}{\arg\min} \left( \sum_{i: x_i \in R_1(j,s)} \left( y_i - \hat{y}_{R_1} \right)^2 + \sum_{i: x_i \in R_2(j,s)} \left( y_i - \hat{y}_{R_2} \right)^2 \right), j = 1, 2, ..., J, \qquad (26)$$

where $R_1$ is the region of the predictor space in which $X_j$ has values less than $s$ and $R_2$ is the remaining region of the predictor space that holds all values of $X_j$ that are greater than or equal to $s$.

After the initial partition, the entire predictor space splits into two regions: $R_1$ represented by child node 1 and $R_2$ represented by child node 2. The same greedy approach is used to further split each of previously formed regions i.e. $R_1$ and $R_2$. After splitting child nodes 1 and 2, the predictor space divides into four regions i.e. regions $R_1$, $R_2$, $R_3$ and the region represented by child node 3. When child node 3 is divided, the resulting regions together with previously formed regions cover the entire predictor space and this completes construction of the tree, hence, the predictive model. For a given test observation, its predicted output is mean of the response values of training observations falling into the region to which the test observation belongs to. Therefore, for each observation belonging to one region, the same prediction is made. The depth of the tree is controlled by minimum size of the leaf nodes, which can be determined using k-fold cross validation approach.

**2.4.2. Bagging regression trees**

As far as prediction accuracy is concerned, a single regression tree (either full tree or pruned tree) is not competitive with regression models developed using ANNs, and suffers from high variance. Bootstrap aggregating (bagging), a general-purpose method, is often exploited, particularly in the framework of decision trees, to reduce variance and improve prediction accuracy of a single prediction model [18]. In bagging, B bootstrapped training datasets from the original training dataset are generated and a separate regression tree ($f^b(x)$) is grown using each bootstrapped training dataset. For a given test observation, output is estimated by taking average ($f_{bagging}(x)$) of the predictions acquired from the B individual regression trees:

$$f_{bagging}(x) = \frac{1}{B}\sum_{b=1}^{B} f^b(x). \tag{27}$$

Each of the B regression trees is grown deep and not pruned, so, it has high variance and low bias i.e. weak learner. Aggregating several weak learners produces a strong learner with high generalization performance, hence, considering a large value of B does not lead to overfitting.

Bootstrap uses only around two-third of the training observations to generate a single dataset and the remaining one-third of the training observations are referred to as out-of-bag (OOB) observations, which can be used to estimate the test error or optimum number of weak learners of the bagged model given in Eq. (27) without performing K-fold cross-validation. For each of $N$ training observations, the response value can be predicted by averaging predictions from B/3 bagged trees in which the observation is OOB. Since the response values of the training observations are known a priori, OOB mean squared error can be computed to get an estimated test error as well as to determine an optimum number of bagged trees.

**2.4.3. Random Forest**

Bagging sometimes cannot yield a significant improvement in prediction accuracy over a single tree. This happens when there is a strong predictor, along with other moderately strong predictors, in the training dataset. This strong predictor is likely to be considered as best split at the root node of most of the trees in the bagged model. As a result, there exists a strong correlation among the trees (weak learners). For a given test observation, predictions gained from the correlated weak learners are virtually the same, hence, average of those similar predictions gives the same value. In this case, no benefit can be achieved from bagging. To overcome this problem, random forest (RF) was proposed [18]. Similar to bagging, RF generates multiple bootstrapped training datasets from the actual training dataset, and fits a separate tree to each bootstrapped dataset. While growing the trees, only a subset of $m$ predictors from the full set of P predictors is randomly selected to split the root node or any internal node. This approach breaks correlation among the trees in the bagged model; thus, averaging uncorrelated trees (weak learners) leads to a significant improvement in prediction accuracy over bagging. There are two parameters in RF model that need to be determined: number of trees (B) and the predictor subset size $m$. A small value of $m$ is selected when there is a large number of correlated predictors; however, the choice is problem dependent.

### 2.4.4. Boosting regression trees

Boosting is another general-purpose method producing an ensemble tree, which is also able to provide more accurate predictions than a single regression tree. The boosting method combines B regression trees (weak learners) into a single prediction model (strong leaner) in an iterative manner. Algorithm I explains different stages involved in the development of the boosted ensemble model. The algorithm starts by considering a constant function for the initial tree. The first regression tree is grown using the modified training data $\{(x_i, r_i)\}_{i=1}^{N}$ i.e. the tree attempts to reduce the errors of the previously developed tree (initial tree). The initial and first tree are added as per Eq. (28) to get final prediction model in the first iteration. In the second iteration, the pseudo-residuals of the final prediction model obtained in the first iteration are computed and used to train $F_2(x)$. The final prediction model obtained in the previous iteration is updated with $F_2(x)$ using Eq. (28). The process is repeated until B trees are developed. Unlike bagging and RF, the boosting method uses weighted sum to combine all weak learners. In general, each of the weak learners possesses a few terminal nodes, thus, the boosting method slowly learns the underlying relationship between the input variables and the target variable. The learning rate (or shrinkage parameter) $\gamma$ further slows down the learning process. Not like bagging and RF, the boosting method leads to overfitting if very large value of B is used, therefore, B needs to be carefully determined along with the number of terminal nodes of each weak learner and the learning rate $\gamma$.

**Algorithm I**

**Given:** Training dataset $\{(x_i, y_i)\}_{i=1}^{N}$, number of weak learners B and learning rate $\gamma$.

**Algorithm**

**Step 1**: Consider an initial tree $F_0(x) = 0$ and set $r_i = y_i$.

**Step 2**: For b=1, 2, …, B, repeat the following

a. Fit a weak learner ($F_b(x)$) to the modified training dataset $\{(x_i, r_i)\}_{i=1}^{N}$.

b. Update the model.
$$F_b(x) = F_{b-1}(x) + \gamma F_b(x) \tag{28}$$

c. Update the residuals.
$$r_i = y_i - F_b(x). \tag{29}$$

**Step 4**: Update the boosted model $F_B(x)$

## 2.5. Kernel methods

### 2.5.1. Gaussian process regression (GPR)

GPR is a non-parametric kernel-based ML method, which is one of popular methods for regression [24]. Gaussian processes (GPs) are nothing but the extension of multivariate Gaussian distributions to infinite dimensionality. As opposed to assuming a specific form for the underlying relation $f(x)$ between the input variables and the response variable, GPR describes $f(x)$ as a multivariate Gaussian distribution, which is fully defined by its mean function $m(x)$ and covariance matrix $K(x, x')$:

$$f(x) \sim GP(m(x), K(x, x')) + \varepsilon, \tag{30}$$

where $m(x) = E[f(x)]$, $K(x, x') = E[(f(x) - m(x))(f(x') - m(x'))]$ and $\varepsilon$ is Gaussian distributed noise with zero mean and variance $\sigma^2$.

It is assumed that the training data are generated from a GP with zero mean. Accordingly, the prior distribution of the output observations is a Gaussian distribution defined below.

$$y \sim N(0, K(x, x')). \tag{31}$$

The covariance matrix $K(x, x')$ is defined as

$$K(x, x') = \begin{bmatrix} k(x_1, x_1) & k(x_1, x_2) & \cdots & k(x_1, x_n) \\ k(x_2, x_1) & k(x_2, x_2) & \cdots & k(x_2, x_n) \\ \vdots & \vdots & \ddots & \vdots \\ k(x_n, x_1) & k(x_n, x_2) & \cdots & k(x_n, x_n) \end{bmatrix}, \tag{32}$$

where $k(x, x')$ is a covariance function parameterized by a set of hyper-parameters.

For a given test data $x_*$, the objective of the GPR model is to predict $y_*$ using the training data. In order to do this, GP assumes that both the training output $y$ and the test output $y_*$ have joint distribution defined below.

$$\begin{bmatrix} y \\ y_* \end{bmatrix} \sim N\left(0, \begin{bmatrix} K(x,x') & k_* \\ k_*^T & k_{**} \end{bmatrix}\right), \qquad (33)$$

where $k_* = [k(x_*, x_1), ..., k(x_*, x_n)]^T$ and $k_{**} = k(x_*, x_*)$.

The actual test output $y_*$ is estimated as the mean $E(y_*)$ of the posterior predictive distribution bestowed in Eq. (34).

$$P(y_* | X, y, x_*) = N(E(y_*), \sigma_*^2), \qquad (34)$$

where $E(y_*) = k_*^T K^{-1} y$ and $\sigma_*^2 = k_{**} - k_*^T K^{-1} k_*$.

The variance of the posterior predictive distribution indicates uncertainty in the estimated output $\hat{y}_*$. It can be noticed from the above discussion that the covariance function plays a key role in obtaining predictions from the GPR model. In the present work, exponential function is chosen as the covariance function:

$$k(x, x') = \sigma_f^2 \exp\left(\frac{\sqrt{(x-x')^T (x-x')}}{\sigma_l^2}\right), \qquad (35)$$

where $\sigma_l$ is the characteristic scale length, and $\sigma_f$ is the signal standard deviation.

The unknown parameters (parameters of Eq. (35) and noise standard deviation) are determined by maximizing the log-likelihood of the training data:

$$J = \underset{\theta}{\arg\max}\left(\log P(y|X)\right) = \underset{\theta}{\arg\max}\left(-\frac{1}{2} y^T K^{-1} y - \frac{1}{2}\log|K| - \frac{N}{2}\log(2\pi)\right). \qquad (36)$$

Eq. (36) is optimized using L-BGFS (limited memory Broyden–Fletcher–Goldfarb–Shanno) algorithm [30].

### 2.5.2. Relevance vector machine

Relevance vector machine (RVM) is a probabilistic Bayesian learning method that utilizes fewer basis functions than a comparable SVM while offering benefits such as probabilistic predictions, automatic estimation of model parameters and ability to use arbitrary basis functions (i.e. basis functions need not satisfy Mercel condition). In RVM, the target vector $y$ is expressed as a sum of its approximation $t$ and noise vector $\varepsilon$:

$$y = t + \varepsilon = \phi w + \varepsilon, \qquad (37)$$

where $w$ is the parameter vector and $\phi$ is the design matrix whose columns include the entire set of $M$ basis vectors.

It is assumed that the noise vector $\varepsilon$ is modelled as Gaussian distribution with zero mean and variance $\sigma^2$ as given in Eq. (38).

$$\varepsilon = N(0, \sigma^2). \tag{38}$$

The parameter $\sigma^2$ is estimated from the training data during the training process.

The likelihood for the target vector is defined as

$$p(t|w, \sigma^2) = (2\pi)^{(-N/2)} \sigma^{-N} \exp\left(\frac{\|t-y\|^2}{2\sigma^2}\right). \tag{39}$$

The prior over parameters, which complements the likelihood in Eq. (39), is given below.

$$p(w|\alpha) = (2\pi)^{-(M/2)} \prod_{m=1}^{M} \alpha_m^{0.5} \exp\left(\frac{-\alpha_m w_m^2}{2}\right), \tag{40}$$

where $\alpha = (\alpha_1, \ldots, \alpha_m)^T$ is hyper-parameter vector, each of which is independently associated with weight. The prior mentioned above imparts sparsity properties to the RVM model.

By merging the likelihood and the prior using Bayes rule, the posterior parameter distribution can be obtained.

$$p(w|t, \alpha, \sigma^2) = \frac{p(t|w, \sigma^2) p(w|\alpha)}{p(t|\alpha, \sigma^2)}. \tag{41}$$

Eq. (41) is a Gaussian distribution with the covariance and mean functions, which are provided below.

$$\Sigma = \left(A + \sigma^{-2} \phi^T \phi\right)^{-1}, \tag{42}$$

$$\mu = \sigma^{-2} \Sigma \phi^T y, \tag{43}$$

where $A = diag(\alpha_1, \ldots, \alpha_M)$.

The hyper-parameters and noise variance ($\sigma^2$) can be obtained by maximizing the logarithm of marginal likelihood with respect to $\alpha$:

$$\begin{aligned} J &= \underset{\alpha}{\arg\min} \left(\log\left(p(t|\alpha, \sigma^2)\right)\right) = \log\left(\int_{-\infty}^{\infty} p(t|w, \sigma^2) p(w|\alpha) dw\right) \\ &= -\frac{1}{2}\left(N\log(2\pi) + \log(|C|) + y^T C^{-1} y\right) \end{aligned}, \tag{44}$$

where $C = \sigma^2 I + \phi A^{-1} \phi^T$.

Sequential sparse Bayesian learning algorithm proposed by Tipping and Faul (2003) is employed to optimize Eq. (44) [31].

Once the optimization algorithm converges, MMP can be estimated for new sample $x^*$ using the following predictive distribution.

$$p(t^*|t,\alpha,\sigma^2) = \int \left( p(t^*|w,\sigma^2) p(w|t,\alpha,\sigma^2) \right) dw. \tag{44}$$

Since the integrand is Gaussian distributed, Eq. (45) is also a Gaussian distribution with mean $\mu^*$ and variance $\sigma_*^2$:

$$p(t^*|t,\alpha,\sigma^2) = N(t^*|\mu^*,\sigma_*^2), \tag{45}$$

where $\mu^* = \mu^T \phi(x^*)$ and $\sigma_x^2 = \sigma^2 + \phi(x^*)^T \Sigma \phi(x^*)$.

The predictive mean $\mu^*$ is the estimated MMP for the test sample $x^*$. The first term of the predictive variance $\sigma_x^2$ is the estimated noise in the trianing dataset while the second term prodives uncertainty in prediction.

## 3. Practical dataset for impure/pure $N_2$-crude oil MMP

### 3.1. Data collection

The authors in [7-12] carried out an extensive study of $N_2$ based EOR process and suggested that MMP is influenced by reservoir temperature ($T$), heptane plus fraction molecular weight ($MWC_7^+$), mole fraction of volatile ($C_1$ and $N_2$) and intermediate components ($C_2 - C_6$, $CO_2$ and $H_2S$) of reservoir oil. The injected gas may have impurities in it, which also have significant impact on MMP. In order to account for those impurities present in $N_2$, average critical temperature ($T_{cm}$) of $N_2$ is considered as the fifth input variable. The authors in [16] created a databank, which covers wide range of MMP and the input variables, by gathering data from reliable literature sources [7-12, 26-28]. The databank contains data for pure and impure $N_2$ based EOR process. The same data bank is adopted in the present work. Statistical characteristics of the data are shown in Table 1. From Table 2, it is seen that the mutual correlation between the input variables is not very significant so that there is no apparent multicolinearity problem. Predictive models based on subspace methods such as principal component analysis, partial least squares, etc. will not yield better estimations than MLR.

**Table 1.** Statistical properties of data

| Variable | Minimum | Maximum | Mean | Standard deviation |
|---|---|---|---|---|
| Tcm (K) | 126.1 | 268.7372 | 165.8122 | 47.3308 |
| Volatile (fraction) | 0 | 0.6055 | 0.3574 | 0.1641 |
| Intermediate (fraction) | 0.1167 | 0.6376 | 0.2335 | 0.0968 |
| $MWC_7^+$ (g/mol) | 140 | 290 | 212.2841 | 46.2212 |
| T (K) | 333.1667 | 464.2167 | 387.3606 | 27.8338 |
| MMP (MPa) | 22.10 | 64.8107 | 37.8111 | 9.4333 |

**Table 2.** Correlation coefficients between input variables

| Input variables | Tcm | Volatile | Intermediate | $MWC_7^+$ | T |
|---|---|---|---|---|---|
| Tcm | 1 | -0.0429 | 0.2162 | 0.3671 | -0.1872 |
| Volatile | -0.0429 | 1 | 0.0378 | -0.1121 | -0.0264 |
| Intermediate | 0.2162 | 0.0378 | 1 | 0.0309 | -0.2794 |
| $MWC_7^+$ | 0.3671 | -0.1121 | 0.0309 | 1 | -0.0252 |
| T | -0.1872 | -0.0264 | -0.2794 | -0.0252 | 1 |

### 3.2. Data partition

Random sampling (or random partitioning) is exceptionally widespread in predictive modelling due to its ease of utilization. However, training set thus created is less likely to be representative of the actual input space of historical data. In addition, it is not ensured that samples on the boundaries of the input space are included in the training set. It is highly probable that models constructed using training sets obtained from the random partitioning method will exhibit poor generalization performance on new unseen samples. Therefore, the present work adopts data partition algorithm (HSPXY) proposed in Li et al. (2019). [35].

The HSPXY algorithm relies on hybrid cosine angle distance and Euclidean distance to select samples for the training set. The cosine angle distance serves as a correlation measure, between a pair of samples, to compensate intra-class relationship:

$$d_{\cos(r,t)} = \frac{\sum_{i=1}^{P} (x_r(i) x_t(i))}{\sqrt{\sum_{i=1}^{P} (x_r(i))^2} \sqrt{\sum_{i=1}^{P} (x_t(i))^2}}, \quad (46)$$

where $x_r(i)$ and $x_t(i)$ are the i$^{th}$ feature of samples $r$ and $t$.

Euclidean distance is to account for the diversity of samples. The HSPXY algorithm prevents correlated samples from being included in the training set i.e. avoiding redudant samples. The hybrid distance measure is defined as

$$D_{xy}(r,t) = 1 + \underbrace{\frac{d_x(r,t)}{\max_{r,t \in [1,N]} d_x(r,t)}}_{A} - \underbrace{\frac{d_{\cos}(r,t)}{\max_{r,t \in [1,N]} d_{\cos}(r,t)}}_{B} + \underbrace{\frac{d_y(r,t)}{\max_{r,t \in [1,N]} d_y(r,t)}}_{C}, \quad (47)$$

where $A$ and $B$ are, respectively, the normalized Euclidean and cosine angle distances in the input space, and $C$ is the normaized Euclidean distance in the output space. The HSPXY algorithm is outlined in Fig. 6.

```
Given size (Q) of training set T and labelled historical database (H).
D_xy for each pair of samples in the historical database is computed. The pair with the largest D_xy
is included in T and removed from H.
For i=3:1:Q
    For n=1:1: number of samples already included in T
        For k=1:1: number of remaining samples in H
            Compute D_xy between n^th sample in T and k^th sample in H:
                D(k) = D_xy(n,k)
        End
        Find sample with minimum D_xy:
        [min, idx_1] = min(D)
        S(n) = H(idx_1,:)
    End
    [max, idx_2] = max(S)
    T(i) = H(idx_2,:)
End
```

**Figure 6**. The HSPXY algorithm.

## 4. Results and discussions

In this section, predictive models using the chosen SL and ML methods will be constructed, evaluated and compared with the predictive models [16, 17] and correlation models [7-12] reported in literature. The data bank was partitioned into training and test datasets using the HSPXY algorithm. The training dataset contains 80% of the databank, and the test dataset holds the rest. For the WNN model, min-max normalization was used to make the input variables and the target variable to be within the range [1, -1]. For the remaining models, the input variables and the target variable were normalized to zero mean and unit variance using z-score normalization method. The performance indices, coefficient of determination ($R^2$) and root mean squared error (RMSE), are used to quantify the generalization performance of the developed predictive models.

### 4.1. Development of predictive models

#### 4.1.1. MLR

MLR model was trained with the training dataset using the method of least squares. The optimum values of the unknow parameters of the model are provided in Table 3.

**Table 3.** Optimal values for MLR parameters

| Parameter | $\beta_0$ | $\beta_1$ | $\beta_2$ | $\beta_3$ | $\beta_4$ | $\beta_5$ |
|---|---|---|---|---|---|---|
| value | $-2.319*10^{-16}$ | -0.3118 | -0.1860 | -0.4514 | 0.0232 | 0.1383 |

### 4.1.2. PR

Table 4 contains the optimum values of the model parameters.

**Table 4.** Optimal values for MLR parameters

| Parameter | Value | Parameter | Value | Parameter | Value |
|---|---|---|---|---|---|
| $\beta_0$ | 0.1188 | $\beta_7$ | 0.0009 | $\beta_{14}$ | 0.1764 |
| $\beta_1$ | -0.2719 | $\beta_8$ | 0.1943 | $\beta_{15}$ | 0.0477 |
| $\beta_2$ | -0.0543 | $\beta_9$ | -0.0564 | $\beta_{16}$ | -0.3770 |
| $\beta_3$ | -0.2868 | $\beta_{10}$ | -0.1224 | $\beta_{17}$ | 0.2960 |
| $\beta_4$ | 0.52622 | $\beta_{11}$ | 0.4479 | $\beta_{18}$ | -0.0362 |
| $\beta_5$ | -0.0383 | $\beta_{12}$ | -0.0389 | $\beta_{19}$ | 0.1326 |
| $\beta_6$ | 0.17648 | $\beta_{13}$ | 0.1755 | $\beta_{20}$ | -0.0685 |

### 4.1.3. WKNN

The number of nearest neighbors was varied in the range [1, 15] with increment of 1, and the 5-fold cross validation approach described in Fig. 1 was applied to the training dataset. In each fold, mean sqaured error (MSE) was computed to quantify the performance of WKNN. For each value of k (i.e. number of nearest neighbors), the mean of MSEs obtained in all the five folds was calculated and plotted in Fig. 7. The smallest MSE was attained when 2 nearest neighbors were used. The WKNN model with k=2 will be employed to make predictions on the test dataset.

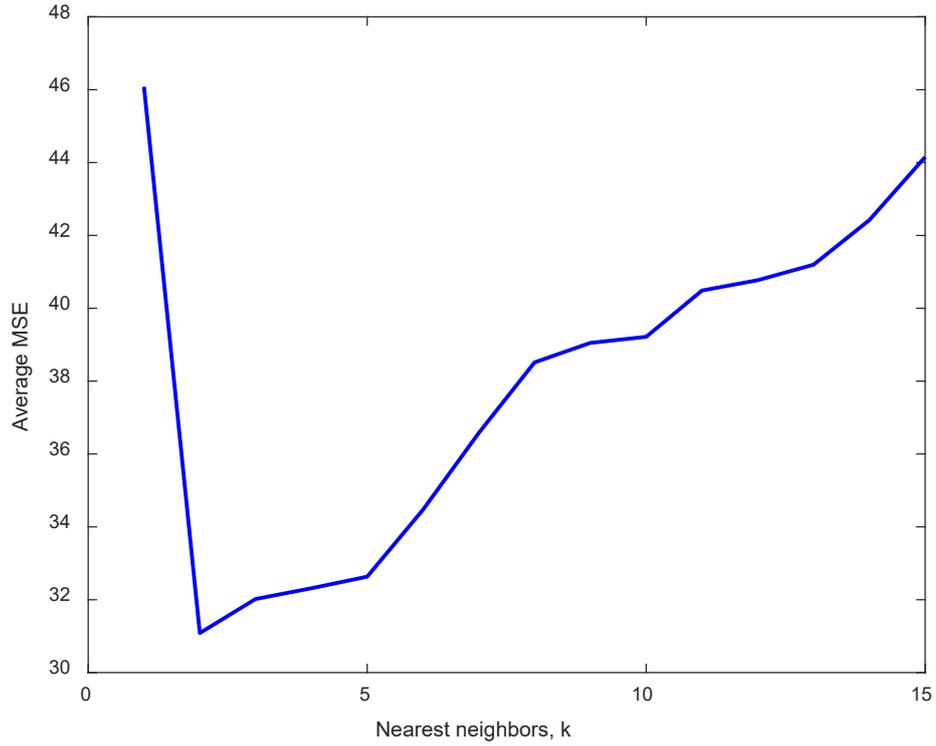

**Figure 7.** Results of the 5-fold cross validation approach for WKNN.

### 4.1.4. WNN

By using Widrow's rule of thumb (given in Eq. (16)) with $N = 67$ and $\varepsilon = 3.5$, the optimum number of neurons in the hidden layer of WNN was determined to be 29. The constants of the B-spline wavelet function (BWF) and gradient descent algorithm were chosen as given in Table 5. The performance of the gradient descent algorithm is displayed in Fig. 8. At the beginning of the training, the model slowly learned the underlyig relationship between the target MMP and the input variables. As the training advanced, the model outputs started to come closer to the actual MMP values.

**Table 5.** Constans used in BWF and the gradient descent algorithm

| Constant | $m$ | $F_b$ | $F_c$ | $\eta_1$ | $\eta_2$ | MaxE |
|---|---|---|---|---|---|---|
| value | 1 | 0.5 | 0.5 | 0.01 | 0.001 | 100 |

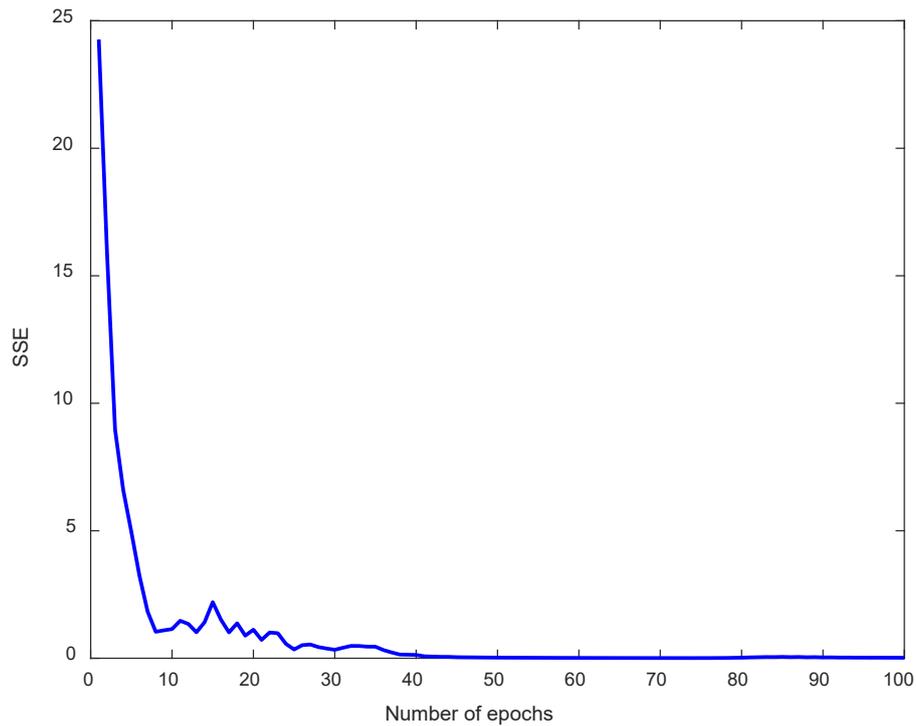

**Figure 8.** Learning curve of WNN.

### 4.1.5. GRNN

As the training dataset contains 67 samples only, the same number of neurons was used in the pattern layer, therefore, the training samples are the centers of the respective radial basiss neurons. The optimum value of the common spread of all the radial basis neurons was determined, via the 5-fold cross validation approach, to be 0.75 as shown in Fig. 9. GRNN was again trained using the optimu spread. As stated in Subsection 2.3.2, GRNN learned all of its parameters in a single pass; hence, there is no learning curve for this network.

### 4.1.6. ELM

Trial and error approach was employed to find the optimum size of the hidden layer of ELM. ELM model with 43 neurons in its hidden layer produced predictions on the training set with resonable accuracy. Similar to GRNN, ELM does not involve iterative training.

### 4.1.7. RT

From the 5-fold cross validation approach (Fig. 10), the optimum size of leaf (or terminal) nodes was found to be 4. A new tree (shown in Fig. 11) was grown using the optimum size of the leaf nodes. In Fig. 11, the leaf nodes are represented by solid dots while the root node and the parent nodes are symbolized by triangles. Each of the leaf nodes is labelled with a constant value i.e. samples falling into the region represented by that leaf node will have that constant as an estimation for MMP. So, RT provides piecewise constant solution for regression problems. The tree shown in Fig. 11 is not symmetric and most of it grew on its right branch.

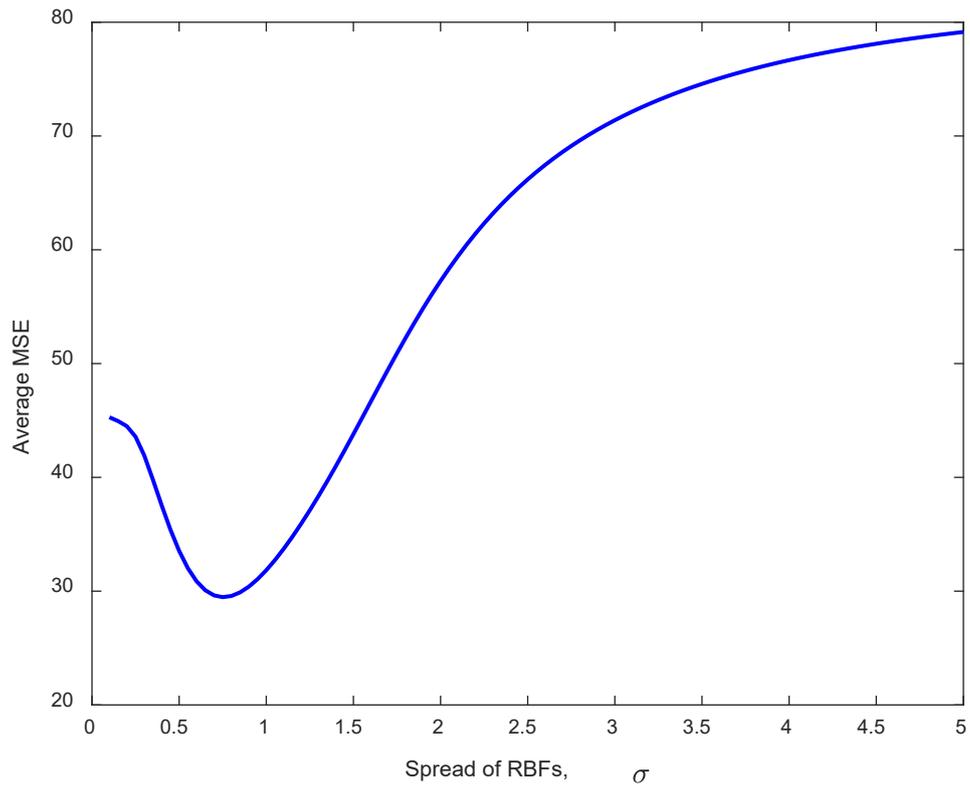

**Figure 9**. Results of the 5-fold cross validation approach for GRNN.

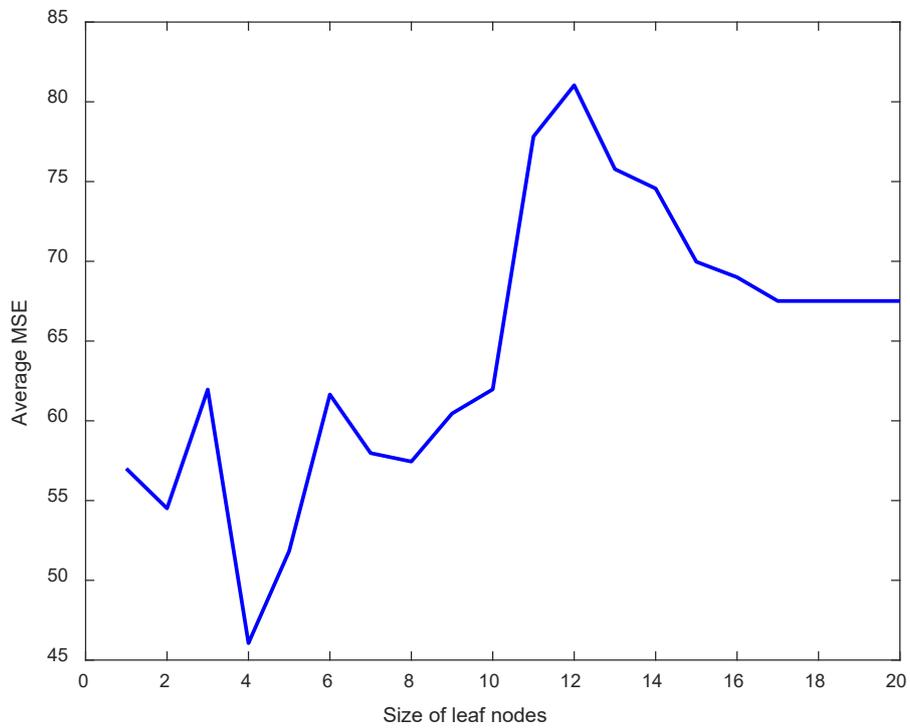

**Figure 10**. Results of the 5-fold cross validation approach for RT

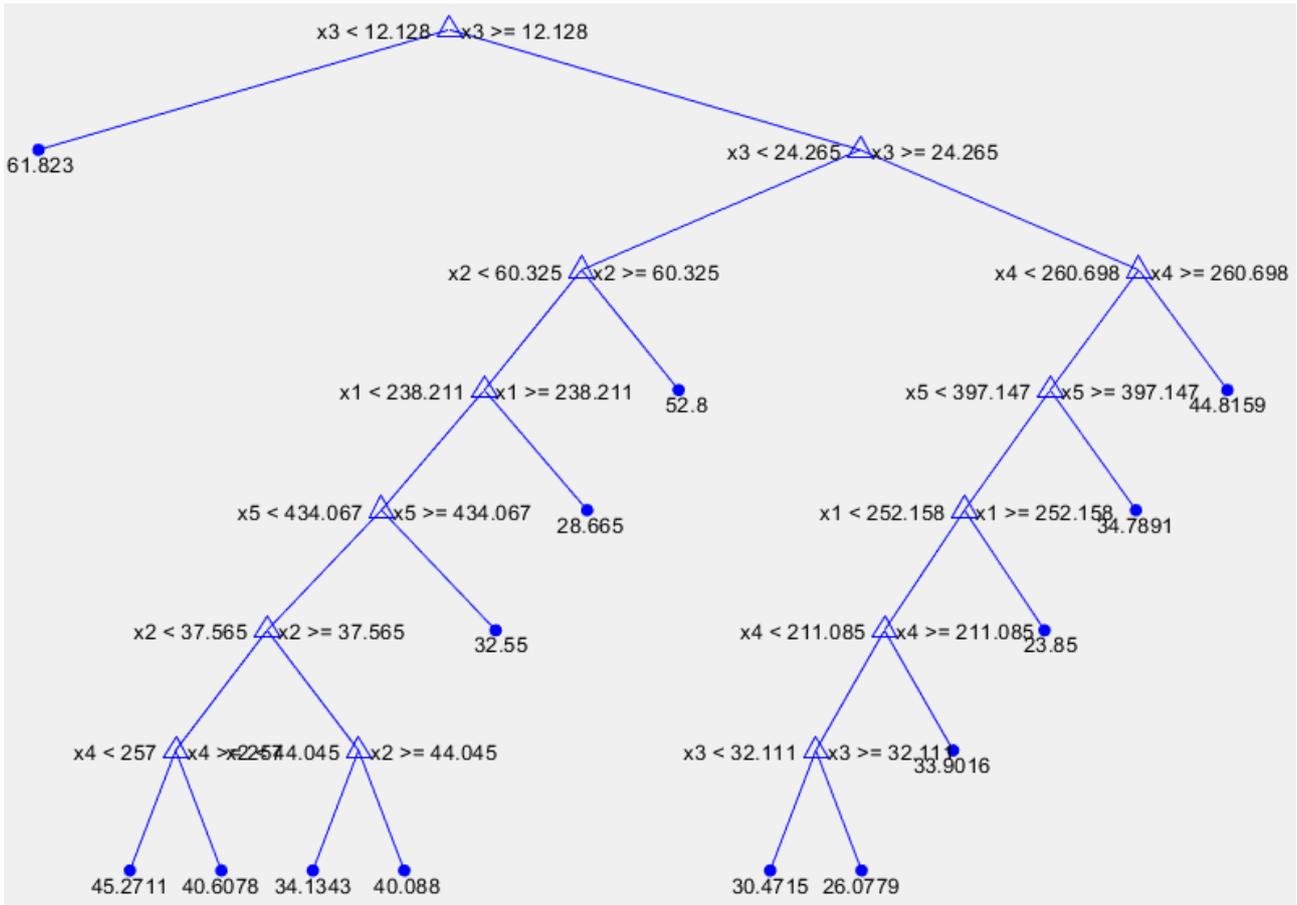

**Figure 11.** Optimally grown tree. In the tree, x1, x2, x3, x4 and x5 indicate Tcm, Volatile, Intermediate, $MWC_7^+$ and T, respectively.

### 4.1.8. Bagging regression trees

In bagging, the weak learners (individual trees) need to be really weak in order for the strong learner (ensemble) to produce better estimations than that can be obtained by any of the weak learners. Therefore, each tree was grown too deep using leaf size of 1. To determine the optimum number of weak learners, ensembles of different sizes were constructed. OOB mean squared error was computed for each of the ensembles built, and plotted against the number of weak learners in Fig. 12. It can be noticed from Fig. 12 that the ensemble with 19 weak learners yielded smallest OOB mean sqaured error of 6.9216.

### 4.1.9. Random forest

In RF, there are two parameters (size of random forest and subset (m) of predictors) that need to be optimally determined. For each value of m in the set {1, 2, 3, 4}, 200 random forests (each random forest is an ensemble or bag of trees with randomly chosen m predictors from the full set of P predictors) were constructed and OOB mean squared error was computed for each of the 200 random forests. According to Fig. 13, the random forest with 26 weak learners and m=4 (subset of 4 predictors) exhibited the smallest OOB mean squared error of 7.3739.

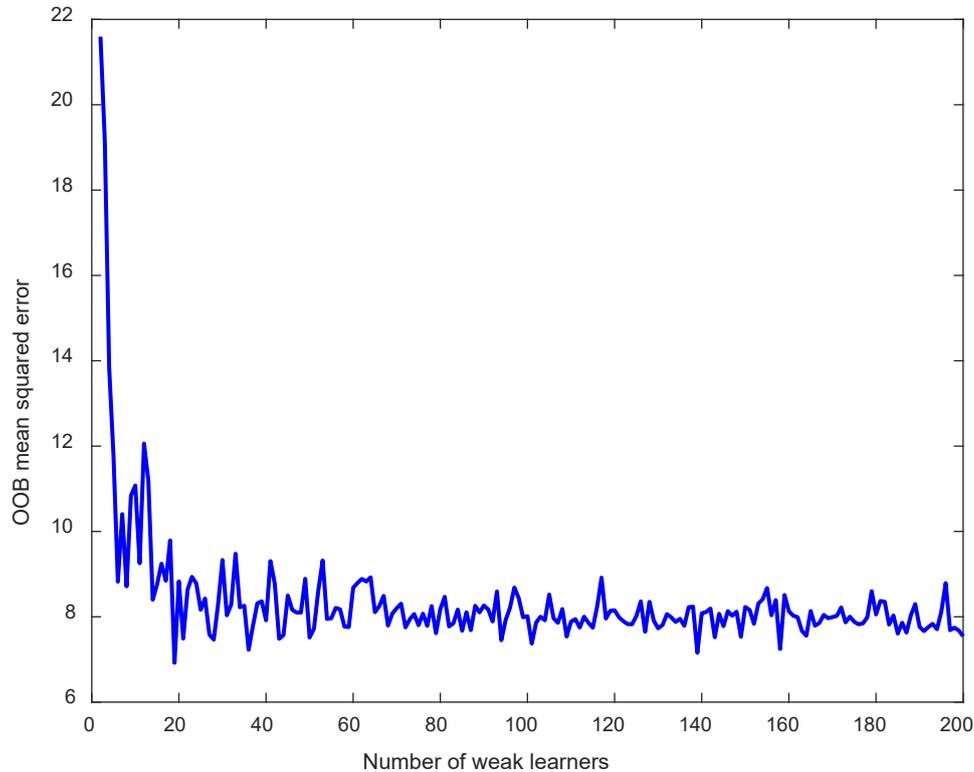

Figure 12. Effect of ensembel size on OOB mean sqaured error.

### 4.1.10. Boosting regression trees

In the boosting regression trees, the ensemble model slowly learns the fundamental relation between MMP and the input variables. Therefore, each weak learner (tree) is not grown too deep, and contains atmost two or three leaf nodes, which are formed by splitting one or two parent nodes, respectively. So, in this work, the maximum number of splits was considered to be 2; hence, each tree has three leaf nodes. As mentioned earlier, unlike the bagging regression trees and the random forest, the ensemble developed from the boosting regression trees approach may encount overfitting problem if a large number of trees is used. The optimum number of trees also depends on learning speed which is describe by learning rate $\gamma$. For different learning speeds, the 5-fold cross validation approach was employed to determine the optimum size of the ensemble. Fig. 14 presents the results of the 5-fold cross validation approach. Based on these results, the optimum size of the ensemble for each value of $\gamma$ was calculated and tabulated in Table 6. The ensemble with NWL=50 and $\gamma$ =0.3 turned out to be the best model.

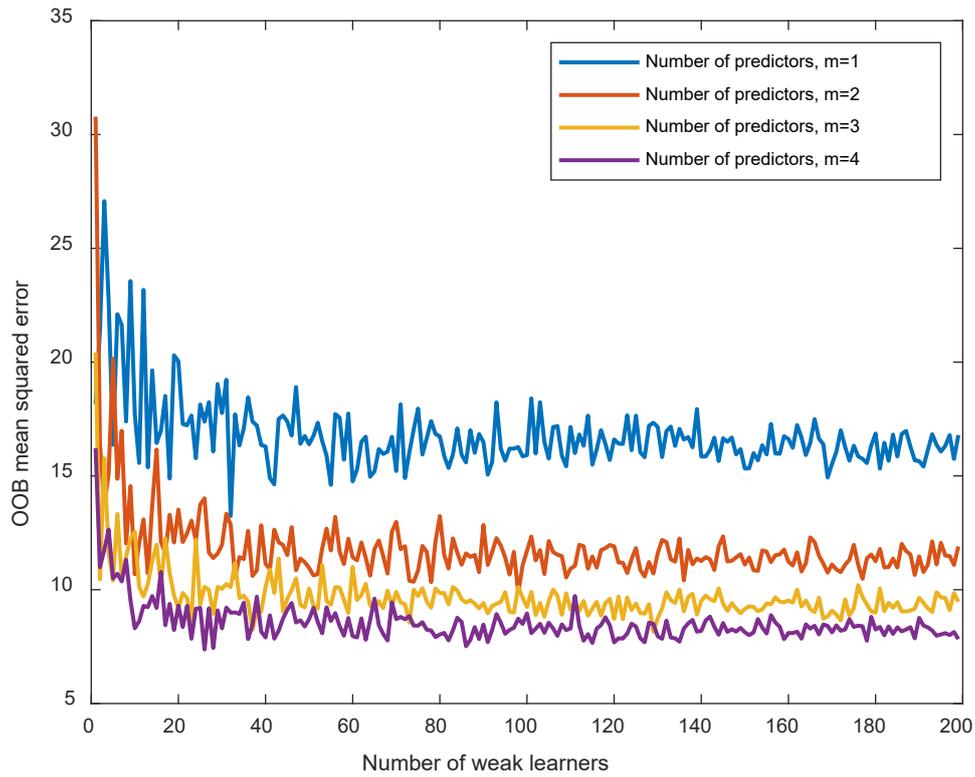

**Figure 13.** Effect of ensembel size and m on OOB sqaured error.

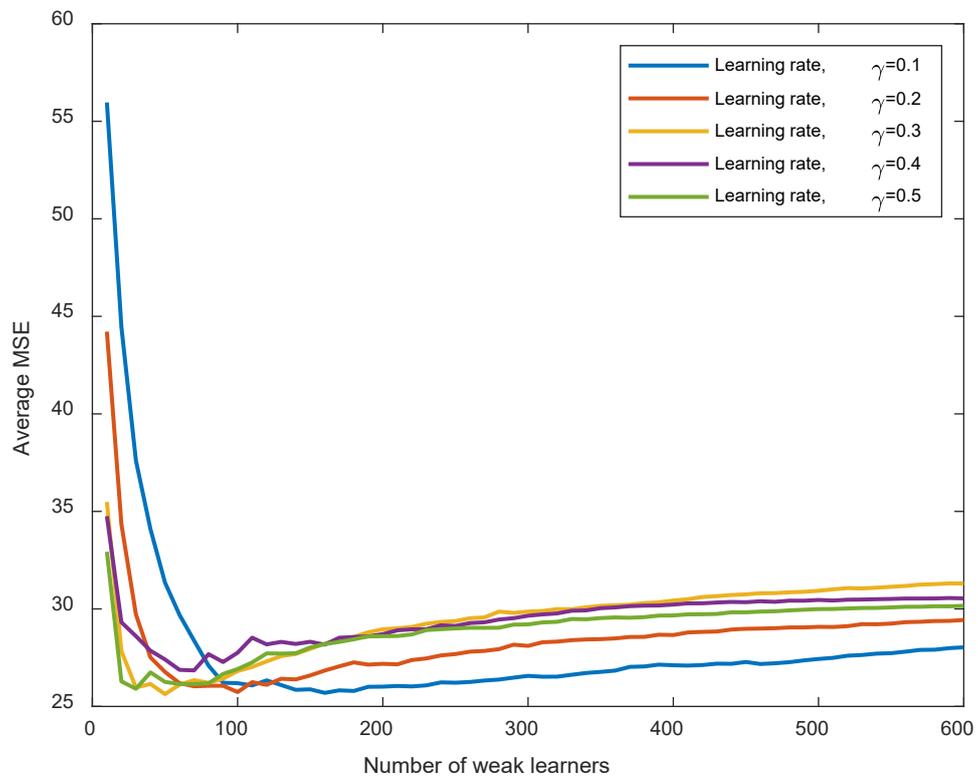

**Figure 14**. Effect of ensembel size and $\gamma$ on average MSE.

**Table 6.** Optimal size of ensemble with respect to learning rate

| Learning rate, γ | 0.1 | 0.2 | 0.3 | 0.4 | 0.5 |
|---|---|---|---|---|---|
| Min AMSE* | 25.6964 | 25.7377 | 25.6322 | 26.8481 | 25.9067 |
| Opimum NWL* | 160 | 100 | 50 | 70 | 30 |

AMSE*-average MSE, NWL*- number of weak learners

### 4.1.11. GPR

Fig. 15 and Table 7 present the results of solving the objective function (log-likelihood) of the GPR model using the L-BFGS optimization method. The L-BFGS method quickly converged the objective function to the local optimum of -65.2856 within 25 iterations.

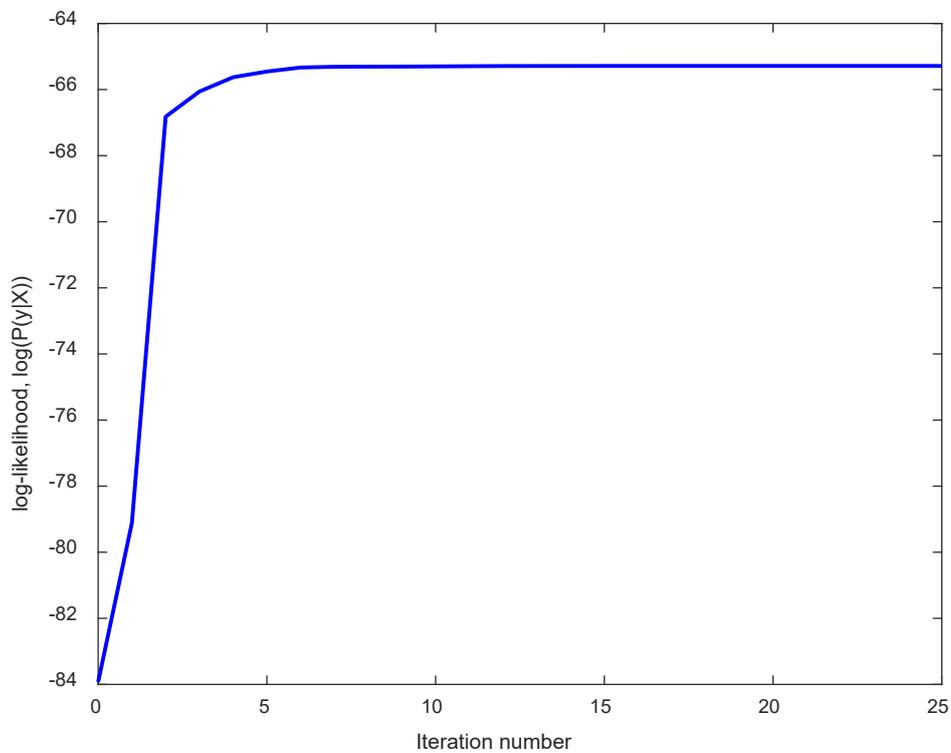

**Figure 15.** Results of the L-BFGS optimization method.

**Table 7**. Hyper-parameters of GPR model

| Value | $\sigma$ | $\sigma_F$ | $\sigma_L$ |
|---|---|---|---|
| Initial guess | 0.7071 | 0.7071 | 1 |
| Optimal value | 0.0227 | 1.3751 | 5.3054 |

### 4.1.12. RVM

The objective function in Eq. (46) was maximised using the sequential sparse Bayesian algorithm and the obtained results are shown in Fig. 16 and Tables 8 and 9. The noise variance was estimated to be 0.6791.

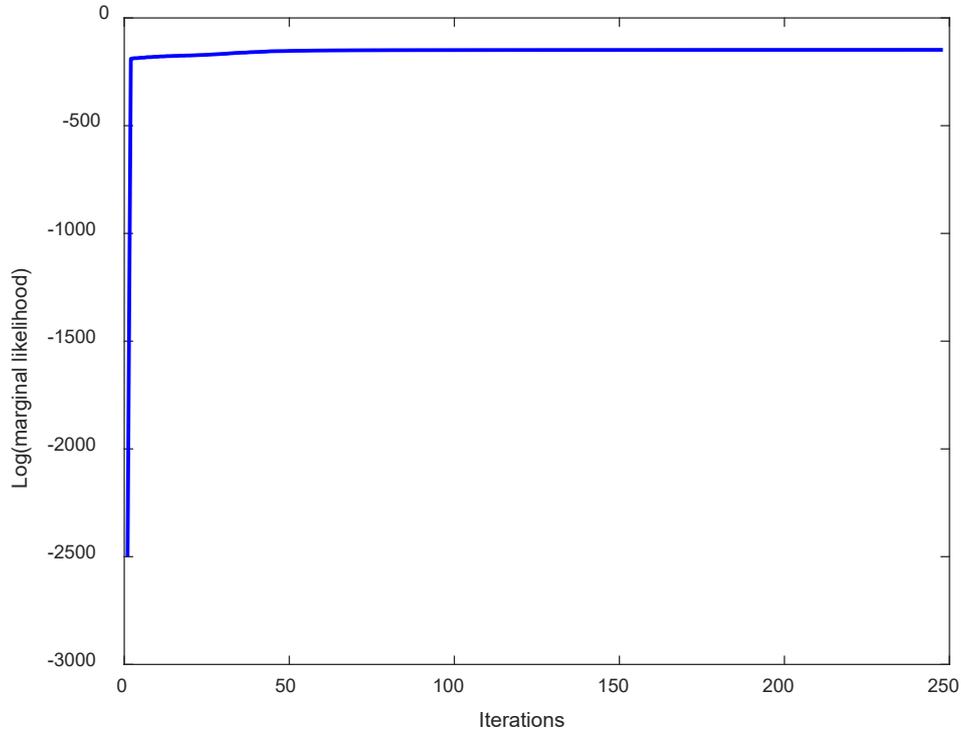

**Figure 16.** Learning curve of RVM model.

**Table 8.** Relevance vector index (RVI) and corresponding weight

| RVI | Weight | RVI | Weight | RVI | Weight | RVI | Weight |
|---|---|---|---|---|---|---|---|
| 1 | 24.1625 | 15 | 2.2280 | 25 | 14.3691 | 33 | 2.0738 |
| 2 | -10.2322 | 16 | -0.4948 | 26 | -0.7925 | 34 | 3.002 |
| 5 | 12.4513 | 18 | 10.1833 | 27 | -0.8296 | 35 | -6.1907 |
| 8 | 6.8718 | 19 | -5.5249 | 28 | -7.5565 | 36 | -1.6655 |
| 11 | 21.4627 | 20 | 6.8667 | 30 | -6.9377 | 39 | 7.9397 |
| 12 | -7.3834 | 21 | 16.9010 | 31 | 25.2692 | 40 | 13.3556 |
| 14 | -8.4674 | 22 | 13.7450 | 32 | 7.7970 | 41 | -4.1917 |

**Table 9.** Relevance vector index and corresponding weight

| RVI | Weight | RVI | | RVI | Weight |
|---|---|---|---|---|---|
| 42 | 4.5142 | 50 | -9.1674 | 60 | 3.8859 |
| 43 | -7.8139 | 52 | 10.5370 | 61 | 18.5930 |
| 44 | 7.8202 | 53 | 1.7175 | 63 | 4.49018 |
| 45 | 9.3187 | 54 | 2.5233 | 64 | 0.16608 |
| 46 | 5.0258 | 56 | -5.6690 | 65 | 4.03539 |
| 47 | -6.4606 | 57 | 4.3585 | 67 | 0.32037 |
| 48 | 10.8735 | 58 | 4.5729 | - | - |

## 4.2. Evaluation of the developed models

The developed predictive models were tested on the test dataset, and the resulting predictions of each of the models are compared with the experimental data in Figs. 17 through 21. The indices ($R^2$ and RMSE) quantifying the generalization performance of the models are provided in Table 10. In the following, the strengths and weaknesses of the SL and ML methods considered in the present study are discussed with respect to the current problem of estimating MMP in the $N_2$ based EOR process.

- Among the chosen predictive modelling methods, MLR demonstrated the poorest prediction performance. The linear nature of MLR prevents it from being able to represent the core relationship between the inputs variables and MMP. The same weakness also reveals that MMP is nonlinearly related to the input variables. Based on this insight, practitioners may search for suitable nonlinear methods rather spending effort, time and resources in searching for linear models.

- Albeit the PR model is linear, the additional terms (interaction and quadratic terms) assisted the model to produce more accurate predictions than MLR. This PR model is only suitable for regression problems with small number of inputs and training samples. The dimensionality of the model will blow up when the number of training sampes and predictors is consierably high.

- Although no learning is involved in WKNN, the obtained predictions are much closer to the experimental data. WKNN outperformed all of the methods expect GPR. However, WKNN also possesses drawbacks: training data storage and exhaustive search. While making predictions, WKNN computes distance between the given test sample and each of the training samples in search of k nearest neighbors. The required storage space and computations will multiply as the size of the training dataset increases. These problems can be dealt with by employing condensed nearest neighbor algorithm to reduce the size of the training dataset.

- Among the three ANNs, the predictions acquired from the WNN model are the most accurate. In comparison to ELM and GRNN, WNN learns more parameters using complex learning algorithm such as the gradient descent based error backpropagation algorithm. Besides, the performance of the gradient descent algorithm depends on the learning rates: $\eta_1$ and $\eta_2$, which are difficult to choose.

- The predictions obtained from the GRNN model are acceptable. The construction of the model is much easier than the WNN model. The only problem with this method is the size of the pattern layer. If $N$ is sufficiently large, it is no longer practical to assign a radial basis neuron to each training sample. In this case, clustering technique like K-means clustering can be employed to group the given training observations, and each group or cluster is represented by a single pattern layer neuron. The mean of the training observations belonging to the $i^{th}$ cluster becomes the center of the $i^{th}$ radial basis neuron.

- The ELM predictions are in accordance with the experimental data and are comparable to that of the WNN model. The attracting feature of ELM is that it has no free parameters, hence, its training speed is several times faster than WNN and GRNN. However, the same feature is also a disadvantage i.e. the performance of ELM cannot be improved beyond a certain level.
- ANNs, GPR and RVM cannot explicitly describe how each input variable is associated with MMP. That is not the case with the decision trees. The decision trees provide a great visual representation of the relation between the input variables and the target variable. The decision trees are easy to construct and understand even by practitioners who has no knowledge in machine learning. On the other hand, a single regression tree cannot provide accurate predictions for MMP. This shortcoming can be overcome by the tree-based ensemble models: bagging RTs, RF and boosting RTs. In the present application, none of the ensembles competes with simpler methods like PR and ELM.
- GPR exhibited the highest generalization performance, and its predictions are virtually equal to the experimental MMP values. Efficient numerical optimization algorithms are needed to learn the hyper-parameters of the model.
- RVM predictions are better than the correlation and predictive models reported in the literature.

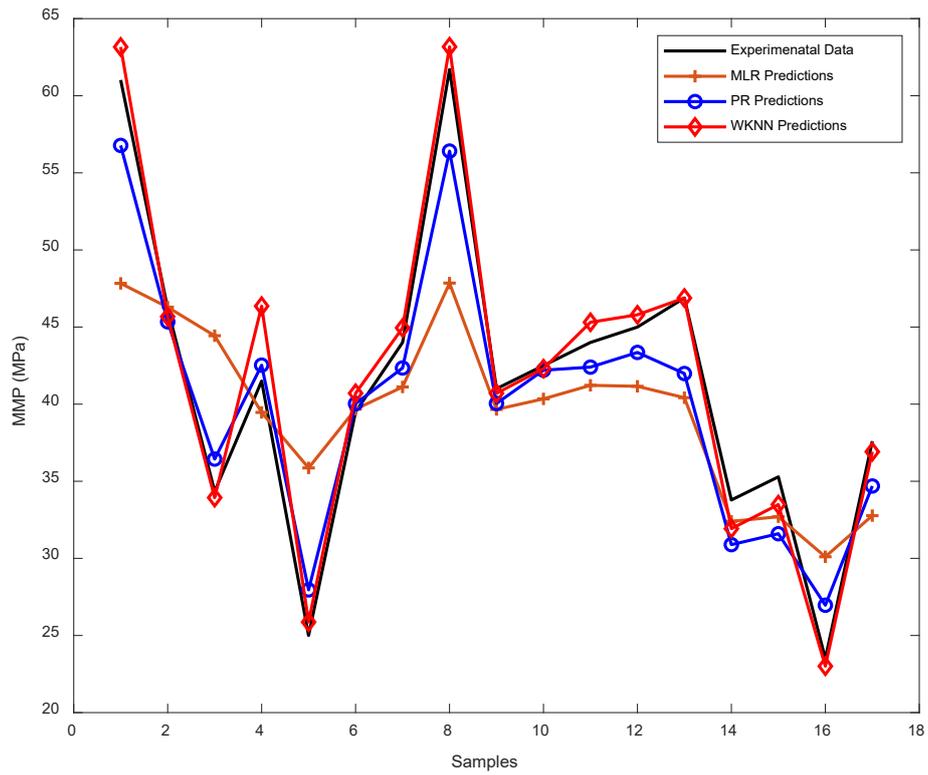

**Figure 17.** Predictions of MLR, PR and WKNN on test dataset

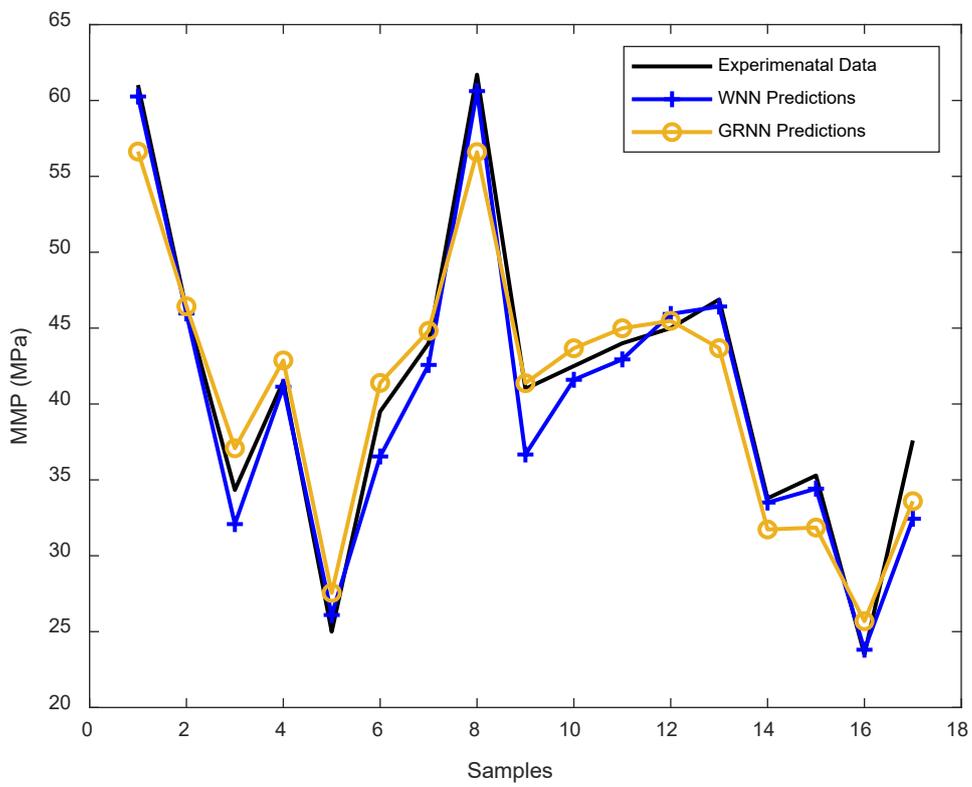

**Figure 18.** Predictions of WNN and GRNN on test dataset

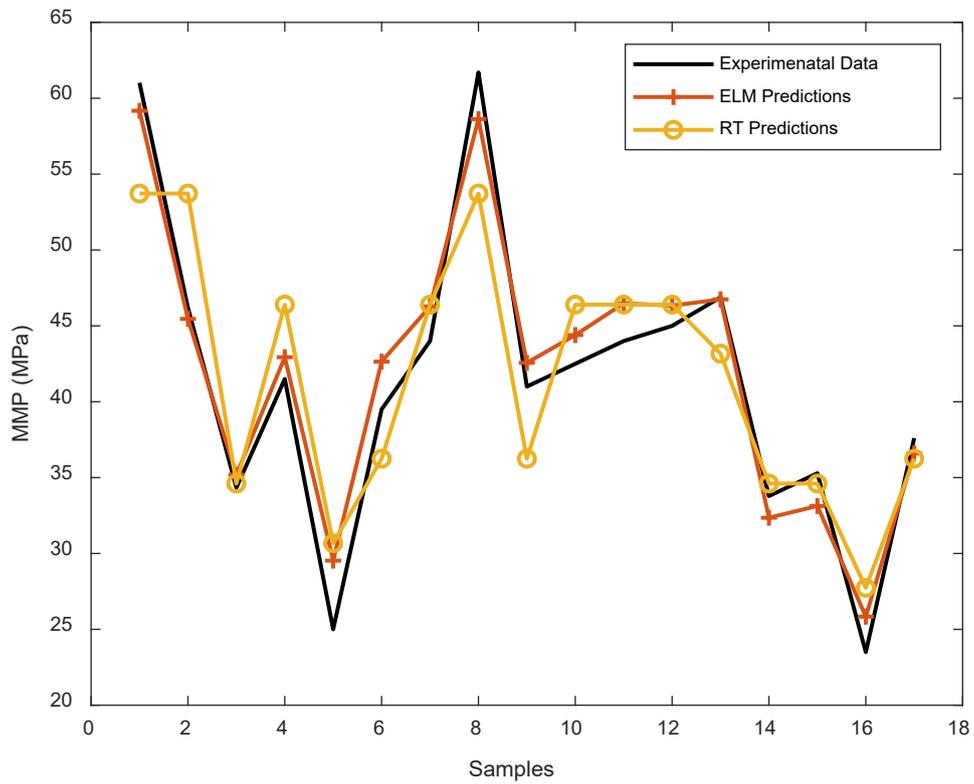

**Figure 19.** Predictions of ELM and RT on test dataset

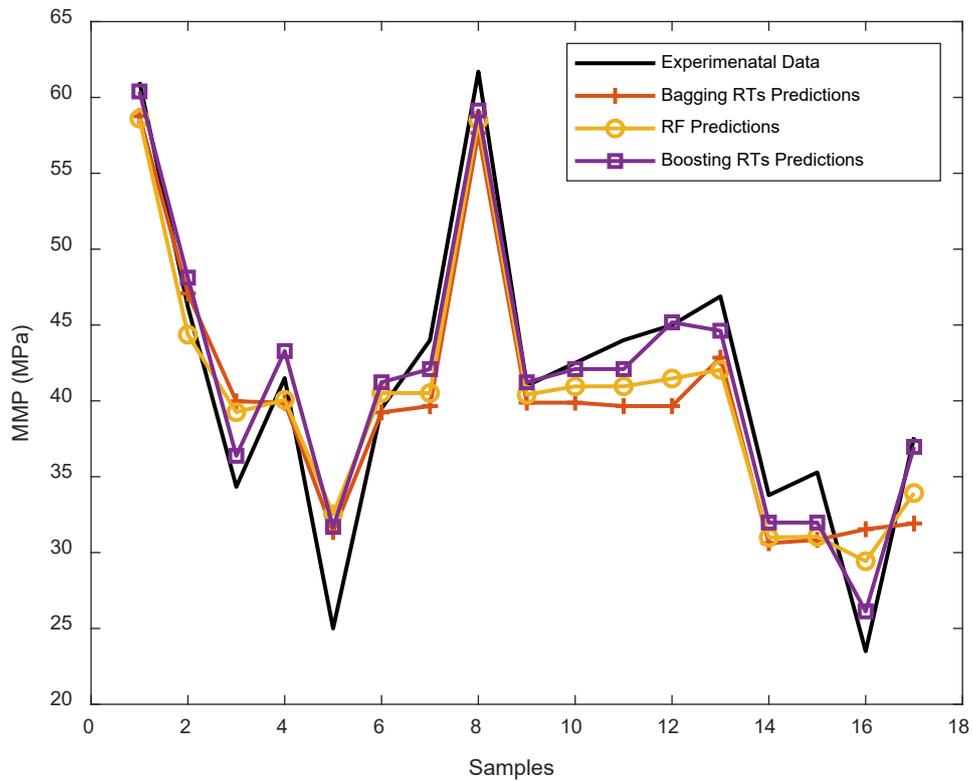

**Figure 20**. Predictions of Bagging RTs, RF and Boosting RTs on test dataset

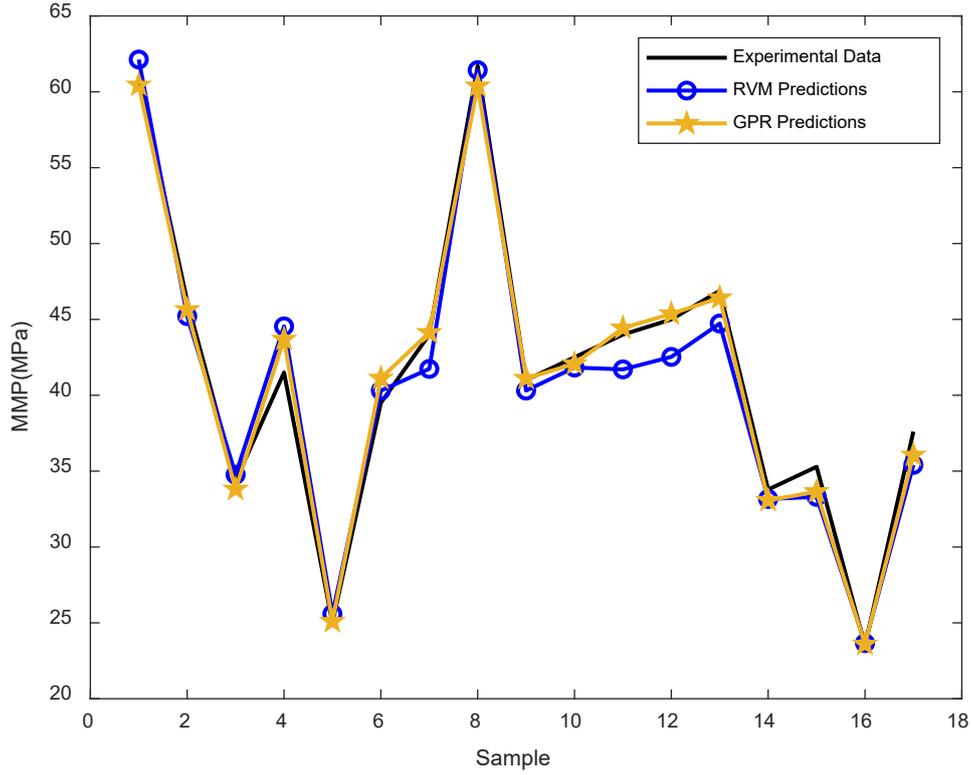

**Figure 21.** Predictions of RVM and GPR on test dataset.

**Table 10.** Generalization performance of models developed in this study

| Model | $R^2$ | RMSE |
| --- | --- | --- |
| MLR | 0.6494 | 6.6181 |
| PR | 0.9530 | 2.8350 |
| WKNN | 0.9817 | 1.6044 |
| WNN | 0.9725 | 2.004 |
| GRNN | 0.9408 | 2.6043 |
| ELM | 0.9621 | 2.1548 |
| RT | 0.8062 | 4.3824 |
| Bagging RTs | 0.8330 | 4.3012 |
| RF | 0.8803 | 3.7398 |
| Boosting RTs | 0.9451 | 2.4241 |
| GPR | 0.9902 | 0.9839 |
| RVM | 0.9772 | 1.6067 |

**Table 11.** Performance indices of models reported in literature

| Model | | $R^2$ | RMSE |
|---|---|---|---|
| Correlation models | Firoozabadi and Khalid [11] | 0.4812 | 8.54 |
| | Hudgins et al. [8] | 0.3146 | 10.27 |
| | Glaso [7] | 0.2387 | 14.70 |
| | Hanssen [9] | 0.3005 | 8.89 |
| | Sebastian and Lawrence [10] | 0.3414 | 9.34 |
| | Fathinasab et al. [12] | 0.7625 | 4.78 |
| Predictive models | LSSVM [16] | 0.9381 | 2.38 |
| | MLP [17] | 0.9186 | 3.31 |
| | RBFNN [17] | 0.9606 | 2.29 |
| | ANFIS [17] | 0.9544 | 2.60 |

## 4.3. Comparision between the present and existing models

Table 11 provides the performance indices of the correlation and predictive models reported in literature for estimating MMP in the $N_2$ based EOR process. In the following, advantages of the models developed in the present work over the existing correlation and predictive models are discussed.

- Except MLR, all other models developed in the present work outperformed the correlation models.
- In [17], genetic algorithm and particle swarm optimization were used to determine the optimal values of parameters of RBFNN and ANFIS models, respectively. The prediction accuracy achieved with either of these models can also be accomplished by WKNN or ELM, which need no or much less training effort, respectively.
- Even if the same learning algorithm was applied to MLP [17] and WNN, WNN predictions are way better than that of MLP due to utilization of wavelets in the hidden layer neurons.
- The linear PR model and the nonlinear GRNN model outpaced the intelligent models: LSSVM and MLP in offering accurate predictions for MMP.
- Only boosted regression trees can compete with LSSVM and MLP because of slow learning. This comparison revelas an important attribute that models learning more slowly will display better extrapolation capability over faster learning models.
- GPR stands out from the models reported here and in literature due to its ability to capture

MMP behaviour on small training dataset. In the case of ANNs and decision trees, sufficient training data are essential.

- Due to sparsity, the RVM predictions are better than that of LSSVM.

## 4.4. Sensitivity analysis

In this section, the effect of the input variables on MMP is evaluated through sensitivity analysis. Two approaches were followed to carry out the sensitivity analysis. In the first approach, the GPR model was supplied with the entire database and Pearson's correlation coefficient between each of the input variables and the resulting estimations of MMP was computed and tabulated in Table 12. Except the reservoir temperature, the remaining inputs are inversely proportional to MMP i.e. if an input variable increases, MMP will decrease. The correlation coefficients obtained with the model outputs are matching the correlation coefficients attained with the experimental data. This comparison shows that the model accurately captured the relation between the input vriables and MMP. In the second approach, the effect of each input variable on MMP was assessed. The first input variable (Tcm) was increased between its minimum and maximum value, and the remaining inputs were kept constant at their minimum values. The predictions obtained from the GPR model on this dataset are shown in the left plot of Fig. 22. In the same fashion, the sensitivity of the remaining input variables was carried out and the results are displayed in Figs. 22 and 24. The results are in good agreement with those obtained in the first approach.

**Table 12.** Correlation coefficients

| Experimental data | Model output |
|---|---|
| -0.4299 | -0.4367 |
| -0.2315 | -0.2233 |
| -0.5606 | -0.5616 |
| -0.0908 | -0.0833 |
| 0.3464 | 0.3487 |

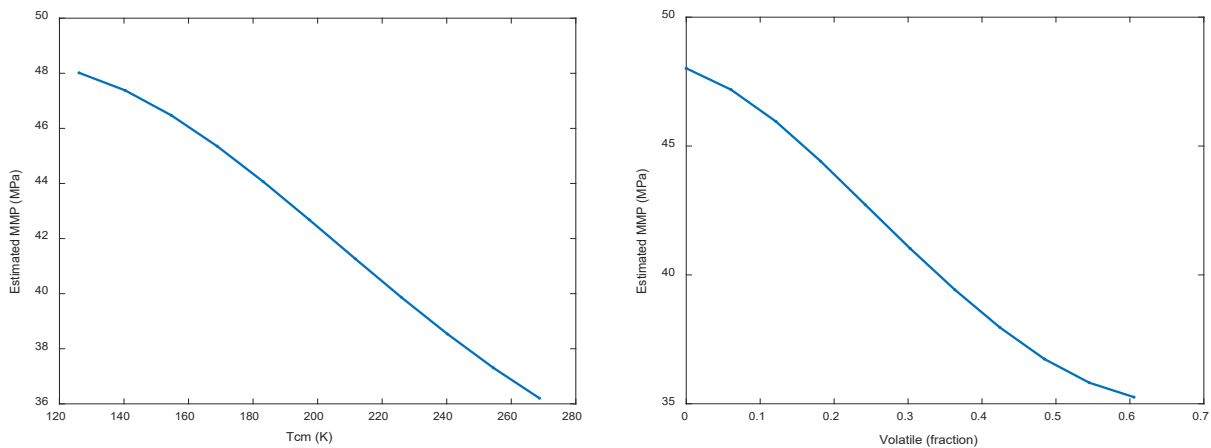

**Figure 22.** Sensitivity analysis of the first and second input variables.

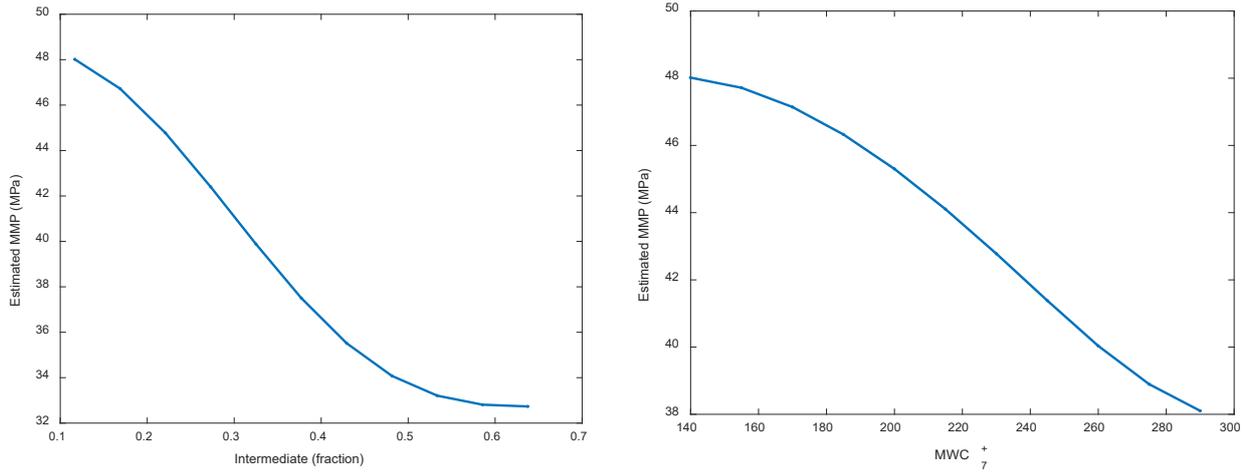

**Figure 23.** Sensitivity analysis of the third and fourth input variables.

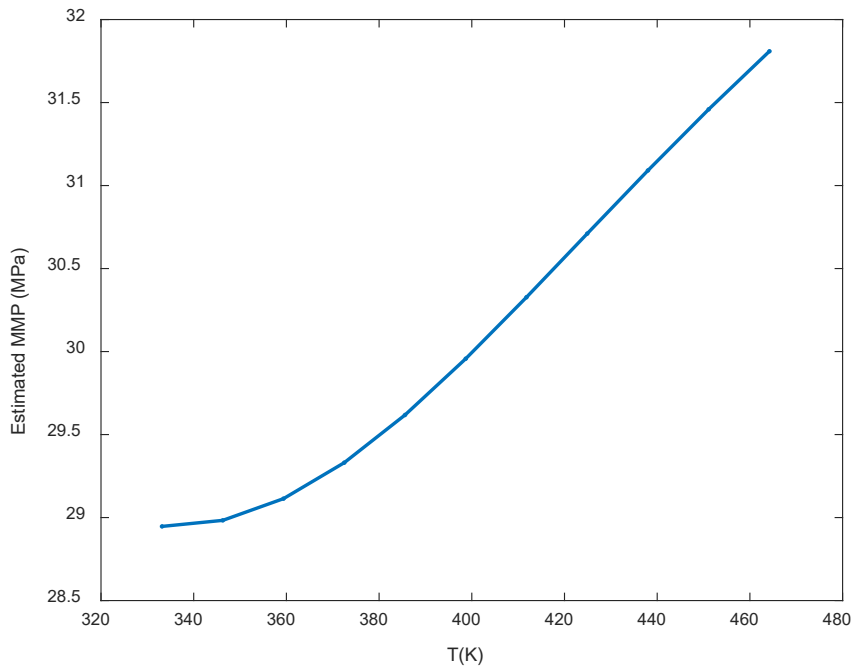

**Figure 24.** Sensitivity analysis of the fifth input variable.

## 5. Conclusions

The successful design of $N_2$ based EOR project greatly depends on the accurate determination of MMP. To overcome the disadvantages of the experimental methods and correlation models (pressure-volume-temperature relationships), researchers resorted to predictive modelling techniques. In this regard, in the present work, predictive models based on different SL and ML methods were developed to estimate MMP of the pure/impure $N_2$-crude oil system. The key findings of the present comparative study are listed below.

- It was proved that the existing correlation models are not highly effective to estimate MMP.

- The present study brought to light elegant ML methods such as WKNN, ELM, RVM and WNN in the context of MMP estimation. These methods were largely overlooked particularly in the research area of predicting the physical and thermodynamic properties of oil and gas.
- The study revealed that when there is a scarcity of training data, kernel methods like GPR and RVM are more useful to construct regression models than ANNs and decision trees.
- Except MLR and the decision trees, the remaining methods demonstrated superior prediction performane over LSSVM, MLP, RBFNN and ANFIS reported in [16, 17].
- The study unveils that like in various areas of science, technology and medicine, ML and AI are becoming indispensable in energy sector to achieve improved performance over traditional practices.

**Acknowledgements**

The authors are grateful to Erfan Mohagheghian, graduate research assistant at University of Calgary, for providing the practical data for pure/impure $N_2$-crude oil MMP.